\pdfoutput=1

\documentclass[11pt]{article}

\usepackage[]{acl}

\usepackage{times}
\usepackage{latexsym}

\usepackage[T1]{fontenc}

\usepackage[utf8]{inputenc}

\usepackage{microtype}
\usepackage{booktabs}
\usepackage{subcaption}
\usepackage{comment}
\usepackage{hyperref}
\usepackage{url}
\usepackage{color,soul}
\usepackage{graphicx}
\usepackage[normalem]{ulem}
\usepackage{mathtools}
\usepackage{algpseudocode}
\usepackage{amsmath,amssymb}
\usepackage[ruled,vlined,linesnumbered]{algorithm2e}

\usepackage{xspace}
\newcommand{\raw}{\texttt{raw}\xspace}

\newcommand{\abttuvar}{\texttt{abtt+zscore}\xspace}
\newcommand{\abtt}{\texttt{abtt}\xspace}
\newcommand{\uvar}{\texttt{zscore}\xspace}
\newcommand{\znorm}{\texttt{zscore}\xspace}

\newcommand{\minmax}{\texttt{minmax}\xspace}
\newcommand{\ulen}{\texttt{ulen}\xspace}

\newcommand{\modelM}{\mathbb{M}}
\newcommand{\dataD}{\mathbb{D}}

\usepackage[colorinlistoftodos,prependcaption,textsize=small]{todonotes}
\presetkeys{todonotes}{inline}{}

\title{Effect of Post-processing on Contextualized Word Representations}

\author{
 Hassan Sajjad$^{\clubsuit}$\thanks{\hspace{1.5mm}The work was done while the author was at QCRI} \hspace{2mm} Firoj Alam$^{\diamondsuit}$ \hspace{2mm} Fahim Dalvi$^{\diamondsuit}$ \hspace{2mm} Nadir Durrani$^{\diamondsuit}$ \\
 $^\clubsuit$Faculty of Computer Science, Dalhousie University, Canada \\
 $^{\diamondsuit}$Qatar Computing Research Institute, Hamad Bin Khalifa University, Qatar \\
 {\sf\small hsajjad@dal.ca,\hspace{1mm}\{fialam,faimaduddin, ndurrani\}@hbku.edu.qa}
}

\begin{document}
\maketitle
\begin{abstract}

Post-processing of static embedding has been shown to improve their performance on both lexical and sequence-level tasks. However, post-processing for contextualized embeddings is an under-studied problem. In this work, we question the usefulness of post-processing for contextualized embeddings obtained from different layers of pre-trained language models.   
More specifically, we standardize individual neuron activations using \emph{z-score}, \emph{min-max} normalization, and by removing top principal components using the \emph{all-but-the-top} method. Additionally, we apply \emph{unit length} normalization to word representations. 
On a diverse set of pre-trained models, we show that post-processing unwraps vital information present in the representations for both lexical tasks (such as word similarity and analogy) and sequence classification tasks. 
Our findings raise interesting points in relation to the research studies that use contextualized representations, and suggest \emph{z-score} normalization as an essential step to consider when using them in an application. 
\end{abstract}

\section{Introduction}
\label{sec:introduction}
Contextualized word embeddings have been central to the recent revolution in NLP, achieving state-of-the-art performance for many tasks. They are commonly used in the form of fine-tuning based transfer learning and feature extraction ~\cite{peters-etal-2018-deep,peters-etal-2019-tune}.\footnote{We have used feature-based transfer learning for this approach in the paper.}
%
%
Feature-based 
approach 
generates contextualized embedding vectors and 
that are used as frozen features in a classifier towards a downstream task.
A similar pipeline is used for static embedding except that here a word 
acquires different embeddings depending on the context it appears in. While fine-tuning based 
is the more commonly used method, feature-based 
approach has shown to be a viable alternative with many applications~\cite{peters-etal-2019-tune}.
For example, it has been used as a tool to analyze the inner learning of pre-trained contextualized models~\cite{dalvi-etal-2017-understanding,liu-etal-2019-linguistic,belinkov-etal-2020-linguistic}.  


The literature on static embedding has emphasized the usefulness of post-processing of embeddings to maximize their performance on the downstream tasks. 
%
For example, \newcite{mu2018allbutthetop} showed that making static embedding isotropic is beneficial to lexical and sentence-level tasks. 
Similarly, \newcite{levy-etal-2015-improving,wilson_2015} showed that using normalized word vectors improve performance on word similarity and word relation tasks.

\begin{figure}[t]
    \centering
\includegraphics[width=0.80\linewidth]{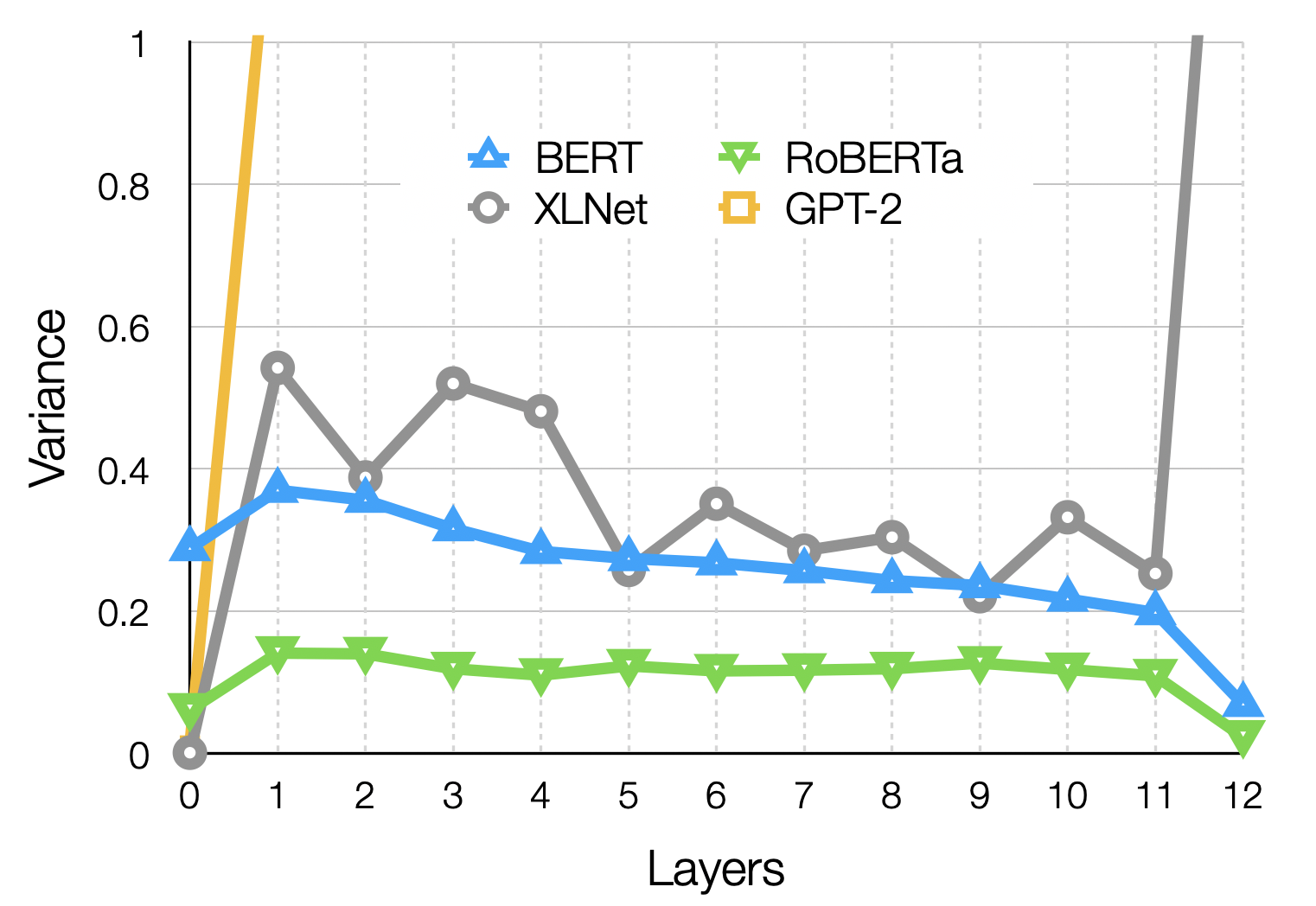}
\caption{Layer-wise variance}
    \label{fig:variance}
\end{figure}

While the efficacy of post-processing has been empirically demonstrated for static embedding, it has not been studied whether it can be beneficial when applied to the representations extracted from the contextualized models such as BERT~\cite{DevlinCLT19}, GPT2~\cite{radford2018improving,radford2019language}, etc. \newcite{ethayarajh-2019-contextual} found contextualized word representations to be anisotropic.
Given that isotropy is a desirable property with proven benefits in the static embedding arena, would encouraging isotropism in the contextualized embeddings also result in performance gains?

Similarly, different layers of a pre-trained model exhibit different variance patterns,
particularly the first and last layers (see Figure~\ref{fig:variance}\footnote{The variance of the middle layers of GPT-2 was very high. We limit the y-axis to show the pattern of variance of other models.} where we plot this for several pre-trained models). When used for feature-based transfer learning, a high variance in features may result in sub-optimal and misleading results~\cite{books/wi/KaufmanR90}. 
Would normalizing the variance in contextualized embeddings be beneficial for their applications? 
It is important to address these questions to unwrap the full potential of contextualized embeddings when used for feature-based learning. In the context of analyzing pre-trained models, they enable a fair comparison between different layers and models. 
%

%

In this paper, we make a pioneering attempt on the realm of post-processing contextual representation. 
To this end, 
we analyze the effect of four post-processing methods on contextualized embeddings. More specifically, we standardize features (individual neurons) using (i) z-score normalization (\uvar), (ii) min-max normalization (\minmax), and (iii) all-but-the-top (\abtt) post-processing method~\cite{mu2018allbutthetop}. We also post-process word representations using unit length normalization (\ulen). The first two are standard feature normalization methods shown to be an effective pre-processing step in building a successful machine learning model~\cite{books/datamining}.   
\abtt removes top principal components of contextualization representations~\cite{ethayarajh-2019-contextual} and improve isotropy of the representations. 
\ulen has shown to be effective in improving the performance of static embedding for word similarity and analogy tasks~\cite{levy-etal-2015-improving}.


We consider contextualized embeddings of a variety of pre-trained models covering both auto-encoder and auto-regressive in design. We evaluate the effect of post-processing contextualized embeddings using various lexical-level tasks such as word similarity, word analogy, and using several sequence classification tasks from the GLUE benchmark.
Our results show that:


\begin{itemize}
\item Post-processing helps to unwrap the 
information present in the representations 

\item The major improvement achieved for the last layer shows that while it is optimized for the objective function, it still possesses most of the knowledge learned in the previous layers 

\item z-score and all-but-the-top are orthogonal and results in best performance when used in tandem for lexical tasks 

\item z-score achieves substantial improvement on the sequence classification tasks, particularly using the representations from the middle and higher layers
\end{itemize}

We further discuss the relation between post-processing of contextualized embeddings and the research on representation analysis. In a preliminary experiment on analyzing individual neurons in pre-trained models, 
we show that post-processing enables a fair comparison among neurons of various layers. For example, after post-processing, we find that the last layer of BERT also has a substantial contribution towards the top neurons  learning part of speech properties. 
%
Supported by our results, we recommend that normalization of a layer representation should be considered as an essential first step before using contextualized embeddings for end applications such as transfer learning and interpretation of representations. 

The paper is organized as follows: Section~\ref{sec:related_work} accounts for related work.
Section~\ref{sec:approach} describes our approach and post-processing strategies. 
Section~\ref{sec:training} presents the experimental setup. Section~\ref{sec:analysis} reports our findings. We carry out the discussion supported by an application in Section~\ref{sec:discussion}. Section~\ref{ssec:conclusion} concludes the paper.

\section{Related Work}
\label{sec:related_work}

\paragraph{Static Embedding Normalization}
A number of post-processing methods have been proposed to improve the performance of static embedding such as, length normalization~\cite{levy-etal-2015-improving}, centering the mean~\cite{sahlgren-etal-2016-gavagai}, and removing the top principal components~\cite{mu2018allbutthetop,arora_2017}. \newcite{arora_2017} removed the first principal component where components are dataset specific as they compute the representations for the entire dataset. On the other hand \cite{mu2018allbutthetop} removed the top $k$ components by computing such representations on the entire vocabulary. They assume that higher variance components are corrupted by the information which is different than lexical semantics.  \newcite{Wang8784743} proposed two normalization methods {\em (i)} variance normalization -- normalizes the principle components of the pre-trained word vectors, {\em (ii)} dynamic embedding -- learns the orthogonal latent variables from the ordered input sequence. The post-processed static representations are then evaluated on both intrinsic and extrinsic tasks, which demonstrates the effectiveness of these methods. 

\paragraph{Contextualized Embeddings}
In the context of representations of contextual pre-trained models, 
the effectiveness of post-processing methods have not been explored. Most of the work that uses contextualized representations use them without any pre-processing. In this work, we explore the usefulness of two commonly used post-processing methods on the embeddings extracted from pre-trained models. 

~\newcite{peters-etal-2018-deep,peters-etal-2019-tune} used contextualized embeddings in feature-based setting for several sequence classification tasks. Similar to static embedding, the contextualized embeddings are used as input to an LSTM-based classifier. However, a word can have different embedding depending on the context. 
In addition, a plethora of work on the analysis and interpretation of deep models used feature-based 
approach to probe the linguistic information encoded in the representations~\cite{belinkov-etal-2017-evaluating,conneau2018you,liu-etal-2019-linguistic,tenney2019learn,tenney-etal-2019-bert,voita-etal-2019-analyzing,durrani-etal-2019-one,arps_syntax:arxiv22}. The most common approach uses probing classifiers~\cite{ettinger-etal-2016-probing,belinkov:2017:acl,adi2017fine,conneau-etal-2018-cram,hupkes2018visualisation}, where a classifier is trained on a corpus of linguistic annotations using representations from the model under investigation. For example, \citet{liu-etal-2019-linguistic} used this methodology for investigating the representations of contextual word representations on 16 linguistic tasks. 
\newcite{dalvi:2019:AAAI,durrani-2020-individualNeurons} expanded the work on representation analysis\footnote{Please see \cite{belinkov-glass-2019-analysis,neuronSurvey} for comprehensive surveys on representation analysis.} to neuron-level analysis. Similar to probing classifier used in the representation analysis, they used a linear classifier with ElasticNet regularization. Recently, \newcite{sajjad:naacl:2022,dalvi2022discovering} introduced an unsupervised method that clusters contextualized representations of words to analyze the representations.


An orthogonal analysis comparing models and their representations relies on similarities between model representations. \citet{bau:2019:ICLR} used this approach to analyze the role of individual neurons in neural machine translation. \citet{wu:2020:acl} compared representations of several pre-trained models using various similarity methods. 
Another class of work on understanding the contextualized representations looked at the social bias encoded in the representations~\cite{bommasani-etal-2020-interpreting}. 

\newcite{ethayarajh-2019-contextual} provided a different angle to the analysis of contextualized embeddings and explored the geometry of the embedding space. They showed that contextualized embeddings are anistropic and questioned the effectiveness of contextualized representations given the well known benefits of isotropic representations. 

The work of \newcite{ethayarajh-2019-contextual} is in-line with ours where they studied the geometry of contextualized representations. Our work builds on top of their analysis and provides an empirical evidence supporting post-processing of embeddings. In relevance to the above mentioned studies on analyzing representation of deep models, our work has direct implications on their findings. The effect of post-processing is dependent on various factors such as, the model used for feature-based transfer learning, and the goal of the task. We suggest that post-processing particularly the z-score normalization should be considered as an essential step while designing experiments using the contextualized embeddings.

\section{Approach}
\label{sec:approach}


Consider a pre-trained model $\modelM$ with $L$ layers ${l_1, l_2 , ..., l_L}$.\footnote{We consider each transformer block as a layer.} Let $\dataD$ be a dataset composed of  $n_w$ words of interest, $w_1, w_2, ..., w_N$. 
Our model $\modelM$ encodes input tokens depending on their context. We first normalize each contextualized embedding using various post-processing methods. 
For each word $w_i$, we then form a single representation, similar to a static embedding, for every layer in $L$. 
We evaluate the word representation on lexical-level tasks and sequence classification tasks. In the latter, we train a BiLSTM classifier pre-initialized with our processed embedding.


\subsection{Word Representations}
\label{sec:wordrepresentations}
Let $s_{w_i}$ be a sentence containing the word $w_i$. Let $z^{l}_{w_i, s_{w_i}}$ represents the contextual embedding from layer $l$ of model $\modelM$ for the word $w_i$, with the given context $s_{w_i}$, specifically, 
\begin{equation}
	z^{l}_{w_i, s_{w_i}} = \modelM(s_{w_i})[w_i]_{l}  \in \mathcal{R}^d 
\end{equation}
where $\modelM(s_{w_i})$ is a shorthand for the contextual embeddings for all tokens in $s$, from which we pick only one that is relevant to the word of interest $w_i$, and further index into a specific layer $l$. $d$ is the number of features (i.e., size of the hidden dimension in a layer), which is $768$ per layer in the models analyzed in this paper.

In order to form a single representation for each word,\footnote{We form a single representation to limit the number of tokens. In Section~\ref{sec:discussion}, we use contextualized embedding of a word in the application. Our results show that our findings generalize to both static and contextualized embeddings.} we first extract contextual embeddings for each word $w_i$ in at least $C_{min}$ contexts. These contexts are derived from a large corpus of sentences,\footnote{We used Wikipedia dump collected on 3rd February 2020.} and are randomly chosen from all sentences containing the word $w_i$. We also employ an upper limit $C_{max}$ for the maximum number of contexts used for any word, to avoid the dominance of frequent words such as closed class words. In our analysis, $C_{min}$ is set to $50$ and $C_{max}$ is set to $200$. Thus, each word in our dataset $\dataD$ will then have between $C_{min}$ and $C_{max}$  contextual embeddings extracted from the model $\modelM$. We then aggregate these contextual embeddings by mean pooling each dimension:

\begin{equation}
	c^{l}_{w_i} = \frac{1}{n_c} \sum^{n_c} z^{l}_{w_i, s_{w_i}}
\end{equation}

where $C_{min} \leq n_c \leq C_{max}$ is the number of contexts (sentences) the word $w_i$ was present in.


\subsection{Post-processing Methods}
\label{sec:methods}
We perform four kinds of post-processing on the representations $z^{l}_{w_i, s_{w_i}}$ before aggregating them into $c^{l}_{w_i}$.
They are; (i) z-score normalization, (ii) min-max normalization, (iii) unit length, and (iv) all-but-the-top normalization. The former two are common feature normalization methods used in machine learning.\footnote{\url{https://en.wikipedia.org/wiki/Feature_scaling}}
The latter two have shown to be effective in post-processing static embedding.
All of these post-processing methods except unit length 
are applied at the feature-level which is a neuron (single dimension) in our case.
The unit length is applied for each word representation. 

Let $\mathcal{Z}$ represents the set of all word occurrence embeddings $z$'s for all words in the dataset $\dataD$.

\paragraph{z-score Normalization (\znorm)}
Centering and scaling input vectors to have zero-mean and unit-variance is a common pre-processing practice in many machine learning pipelines. Concretely, each feature's (in our case 1 of 768 dimensions from each layer's representation) mean and variance is computed across all words in our dataset $\dataD$, followed by subtraction of the mean and division of the standard deviation for each of the feature's value for each embedding  $z^{l}_{w_i, s_{w_i}}$.

\begin{equation*}
	\mu	_l = \frac{1}{|\mathcal{Z}|} \sum_{z \in \mathcal{Z}} z \in \mathcal{R}^d 
\end{equation*}

\begin{equation*}
	\sigma_l = \sqrt{\frac{1}{|\mathcal{Z}|} \sum_{z \in \mathcal{Z}} (z - \mu_l)^2} \in \mathcal{R}^d 
\end{equation*}

\begin{equation}
		\widetilde{z}^{l}_{w_i, s_{w_i}} = \frac{z^{l}_{w_i, s_{w_i}} - \mu_l}{\sigma_l} \in \mathcal{R}^d 
\end{equation}

where $\widetilde{z}^{l}_{w_i, s_{w_i}}$ is the post-processed representation of word $w_i$ in context $s_{w_i}$ from layer $l$.

\paragraph{Min-max Normalization (\minmax)}
The min-max normalization rescales each feature range between 0 and 1. Given values of a feature, \minmax is calculated as follows:

\begin{equation}
		\widetilde{z}^{l}_{w_i, s_{w_i}} = \frac{z^{l}_{w_i, s_{w_i}} - \min(z)}{\max(z) - \min(z)} \in \mathcal{R}^d 
\end{equation}

where $\min$ and $\max$ represent the minimum and maximum values of feature $z$.

\paragraph{Unit Length (\ulen)}
The normalization of word vectors to unit length is shown to be effective for static embedding~\cite{levy-etal-2015-improving}. Here, we evaluate its efficacy for contextualized embeddings. 
Different from other post-processing methods mentioned in this work where we normalize each feature, \ulen is applied at each word vector i.e., a set of features that represent a vector. We normalize vectors using $L_2$ norm.

\paragraph{All-but-the-top (\abtt)}
\newcite{mu2018allbutthetop} showed that all word representations share a large common vector and similar dominating directions which influence them. Eliminating such directions yields isotropic word representations with better self-normalization properties.

We hypothesize that contextualized embeddings belonging to different layers might be influenced by various design factors e.g., initial layers of BERT may have a strong influence of position embedding, ~\newcite{kovaleva-etal-2019-revealing} showed that the last layer of BERT is optimized for the objective function. These factors if dominating the contextual representations may result in sub-optimal performance. We apply \abtt to eliminate such kind of dominating directions.  

\begin{algorithm}[h]
\small
\SetAlgoLined
\DontPrintSemicolon
\KwIn{Word representations $\{v(z) = z^{l}_{w_i, s_{w_i}}$, $z^{l}_{w_i, s_{w_i}} \in \mathcal{Z}\}$, a threshold parameter $k$}
Compute the mean of $\{v(z), z \in \mathcal{Z}\}, \mu \leftarrow \frac{1}{|\mathcal{Z}|} \sum_{z \in \mathcal{Z}} v(z), \tilde{v}(z) \leftarrow v(z) - \mu $\;
Compute the PCA components: $u_1, ..., u_d \leftarrow PCA(\{\tilde{v}(z), z \in \mathcal{Z}\})$. \;
Preprocess the representations: $v'(z) \leftarrow \tilde{v}(z) - \sum_{i=1}^{k} (u_i^T v(z)) u_i$ \;
\KwOut{Processed representations $\widetilde{z}^{l}_{w_i, s_j} = v'(z)$.}
\caption{All-but-the-top procedure}
\label{algo:abtt}
\end{algorithm}

Algorithm \ref{algo:abtt} shows the overall process of \abtt from the original paper, modified with the notation used in this paper. Here, the number of dominating directions to remove is a hyper-parameter $k$. We used the recommended value of $k$ as suggested by \newcite{mu2018allbutthetop} which is $k=d/100$ where $d=768$.

\section{Training and Evaluation Setup}
\label{sec:training}

\paragraph{Data}
We used the Wikipedia dump of 124 million sentences collected on 3rd February 2020. 
We tokenized the text using the Moses tokenizer~\cite{koehn2007moses}.
Given a pretrained model, we extracted the contextualized embedding of words from the Wikipedia corpus and generated a single word representation as described in Section~\ref{sec:wordrepresentations}. 



\paragraph{Contextualized models}
We analyzed the contextualized embedding of four 12-layer pre-trained models: BERT~\cite{DevlinCLT19}, RoBERTa~\cite{roberta}, XLNet~\cite{yang2019xlnet} and GPT2~\cite{radford2018improving,radford2019language}. The former two are auto-encoder in nature while the latter two are auto-regressive.

\paragraph{Lexical-level tasks}
We used seven word similarity datasets: WordSim353 split into similarity and relatedness~\cite{wordsim,wordsim2}, MEN~\cite{bruni-etal-2012-distributional}, Mechanical Turk~\cite{radinsky_mturk}, Rare Words~\cite{luong-etal-2013-better}, SimLex-999~\cite{hill-etal-2015-simlex} and RG65~\cite{rg65}. The datasets contain a word pair with their human annotated similarity scores. The quality of a word embedding is calculated based on the cosine similarity score between a given pair of words, in comparison with their human-provided scores.

Moreover, we used three analogy datasets: MSR~\cite{mikolov2013linguistic}, Google~\cite{google_analogy} and SemEval2012-2~\cite{semeval2012_2}. The analogy tasks involve predicting a word given an analogy relationship like ``a is to b'' as ``c is to d'' where d is the word to predict. 
For more details on each task, we refer the reader to \newcite{levy-etal-2015-improving}. 
We used the word embedding benchmark toolkit\footnote{\url{https://github.com/kudkudak/word-embeddings-benchmarks}} to evaluate word representations on the word analogy and word similarity tasks. 

\paragraph{Sequence classification tasks}
We evaluated using six General Language Understanding Evaluation (GLUE) tasks ~\citep{wang-etal-2018-glue}:\footnote{See Appendix for data statistics and download link.} 
SST-2 for sentiment analysis with the Stanford sentiment treebank \cite{socher-etal-2013-recursive}, MNLI for natural language inference \cite{williams-etal-2018-broad}, QNLI for Question NLI \cite{rajpurkar-etal-2016-squad}, 
RTE for recognizing textual entailment \cite{Bentivogli09thefifth}, MRPC for Microsoft Research paraphrase corpus \cite{dolan-brockett-2005-automatically}, and STS-B for the semantic textual similarity benchmark \cite{cer-etal-2017-semeval}. We compute statistical significance in performance differences using McNemar test.

We trained a BiLSTM model using Jiant~\cite{phang2020jiant}, with the following parameters settings: vocabulary size 30k, sequence length 150 words, batch size 32, dropout 0.2, hidden layer size 1024, number of layers 2, AMSGRAD, learning rate decay 0.99, minimum learning rate 1e$^{-06}$. The embedding layer is of size 768 for all experiments.

\begin{figure*}[t]
    \centering
    \begin{subfigure}[b]{0.32\linewidth}
    \centering
    \includegraphics[width=\linewidth]{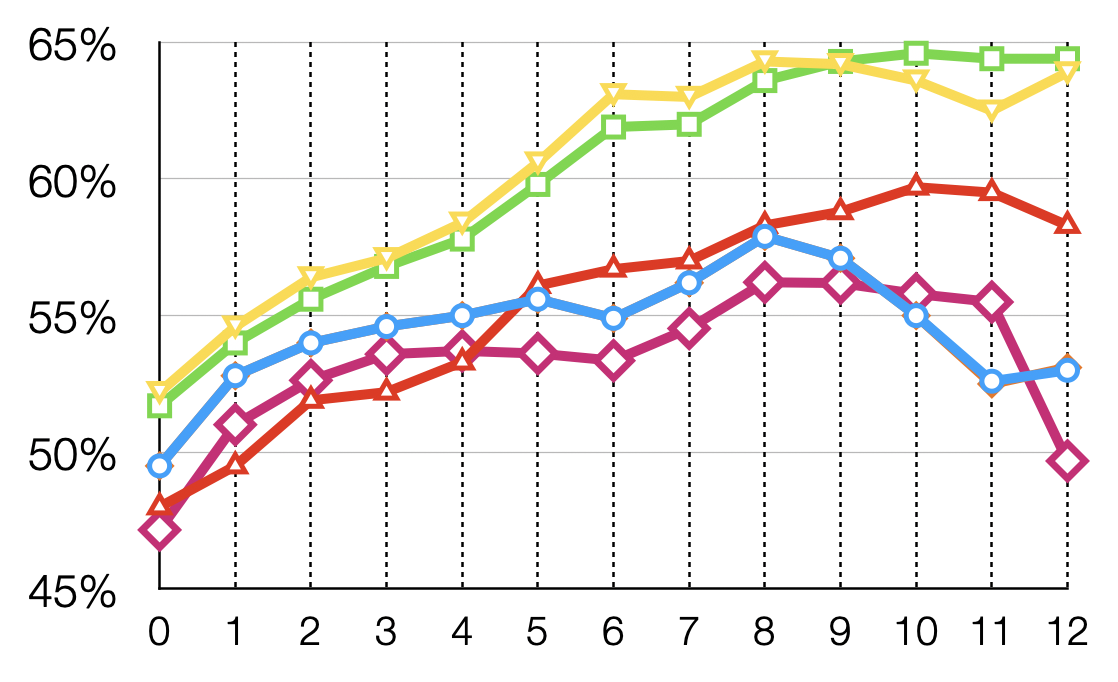}
    \caption{BERT}
    \label{fig:bert_avg_intrinsic}
    \end{subfigure}
    \begin{subfigure}[b]{0.32\linewidth}
    \centering
    \includegraphics[width=\linewidth]{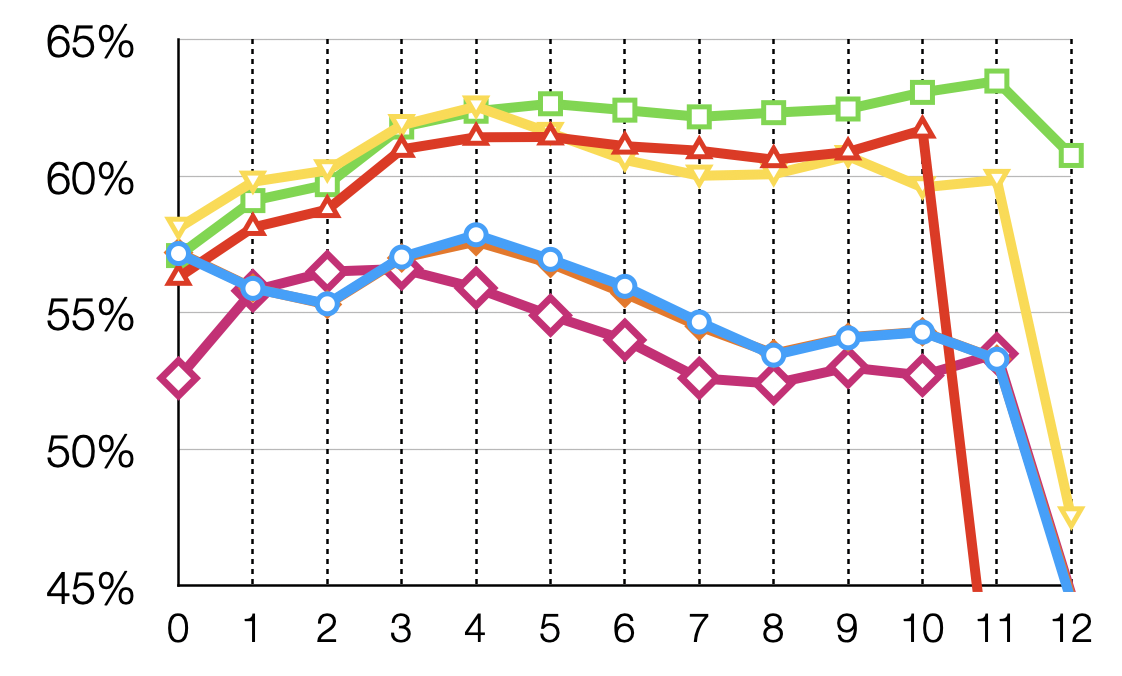}
    \caption{RoBERTa}
    \label{fig:roberta_avg_intrinsic}
    \end{subfigure}    
    \begin{subfigure}[b]{0.32\linewidth}
    \centering
    \includegraphics[width=\linewidth]{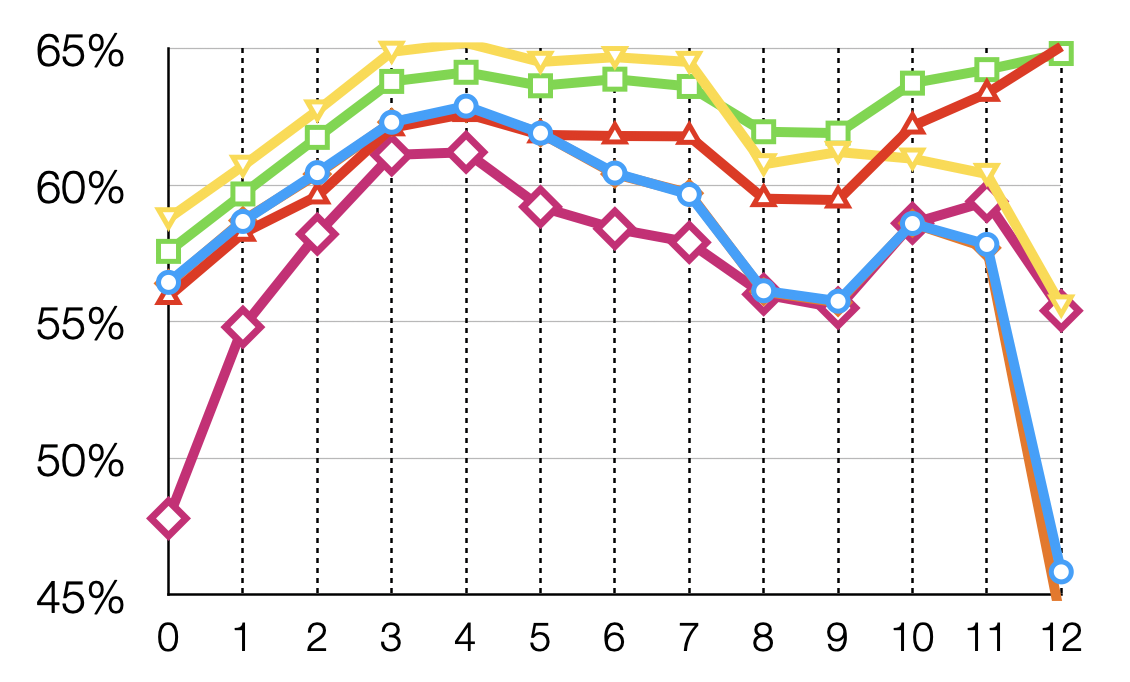}
    \caption{XLNet}
    \label{fig:xlnet_avg_intrinsic}
    \end{subfigure}
    \begin{subfigure}[b]{0.6\linewidth}
    \centering
    \includegraphics[width=\linewidth]{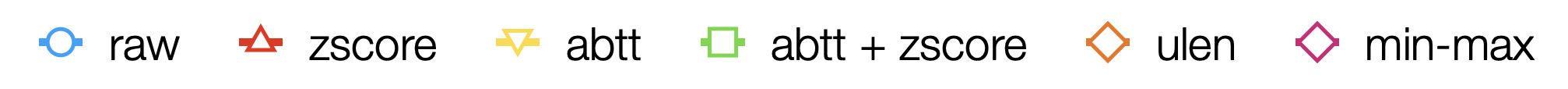}
    \end{subfigure}
    \vspace{-2pt}
    \caption{Average lexical tasks results. X-axis: layers where 0 is the embedding layer}
    \label{fig:avg_intrisic_results}
\vspace{-4pt}
\end{figure*}

\section{Analysis and Findings}
\label{sec:analysis}

We experiment with four post-processing methods as mentioned in Section~\ref{sec:methods}. 
%
We analyze the effect of each post-processing using lexical and sequence classification tasks. Due to limited space, we present the average results of three models only. The task-wise results of all models including GPT2 are provided in Appendix \ref{sec:app_results_intrinsic_tasks} and \ref{sec:app_seq_classification_results}. We did not observe any difference in task-specific trends compared to the average trend present in the paper.

\subsection{Lexical-level Tasks}
Figure \ref{fig:avg_intrisic_results} presents the average layer-wise results using the lexical tasks. \raw represents the embedding before applying any post-processing.

\textbf{Post-processing is generally helpful.} Comparing \raw (blue line) with the rest, other than a few exceptions, layers of all models benefited from the post-processing steps. Surprisingly, \ulen did not result in any change in the performance compared to \raw.\footnote{The results of \ulen and \raw were identical up to two decimal points. The line of \ulen is not visible because it is hidden behind \raw in the figure.} \minmax resulted in poor performance than \raw. 
The two promising post-processing methods are \uvar and \abtt. In the following, we mainly discuss the results of \uvar and \abtt. 

\textbf{Higher layers achieve major improvements in performance.} The general performance trend from lower layers to higher layers suggest that it is essential to post-process the representations of higher layers in order to unwrap the information present in those layers for the lexical tasks. In other words, the higher layer representations though optimized for the objective function still possess similar or better lexical-level information compared to the lower layers. 

\textbf{Comparing post-processing combinations.}
With the exception of the lower layers of BERT and the last two layers of RoBERTa, \uvar (red line) achieved competitive or better performance than using the \raw (blue line) representations. 
Comparing the variance of each layer (Figure \ref{fig:variance}), \znorm is very effective for high variance layers such as the last layer of XLNet and most of the layers of GPT2 (see Appendix~\ref{sec:app_gpt_glue_tasks}). While RoBERTa has the most smooth variance curve, \znorm is still quite effective in improving the performance from layer 1 to 10. The sudden drop in the performance of the last two layers is surprising. The reason could be an extremely low variance of these layers as can be seen in Figure \ref{fig:variance} and applying \znorm alone amplifies the amount of noise.

\abtt outperformed or is competitive to the best performing individual post-processing methods (yellow line in Figure~\ref{fig:avg_intrisic_results}). The consistent improvement across all models for the lower layers 
%
reflects that the lower layer representations consist of top principle components that negatively influence the representations in the context of lexical tasks. For example, the representations from the lower layers of BERT might have a strong influence of position embeddings which may be harmful for word similarity and analogy tasks. On the top two layers, \abtt does not seem to be very effective on RoBERTa and XLNet
suggesting fewer 
dominating principle components in the higher layers. 

Since \uvar and \abtt targets different properties in the representation, we apply them in succession. Using both methods in any order resulted in better representation quality especially for the higher layers (see green line \abttuvar that outperformed or has competitive performance to the best result on all layers). These results show the potential of combining various post-processing methods, like \abttuvar, in achieving better performance on the lexical tasks.

\textbf{Comparing models.}
While post-processing methods benefited all models, XLNet showed the most increase in the performance with lower-middle layers (3,4) and higher layers outperforming all layers across all models. BERT also showed similar gains with more consistent trend i.e. an increase in performance with every higher layer. We did observe gains for RoBERTa. However, they are less substantial compared to BERT and the results on last layer dropped compared to other layers.



\begin{figure*}[t]
    \centering
    \begin{subfigure}[b]{0.31\linewidth}
    \centering
    \includegraphics[width=\linewidth]{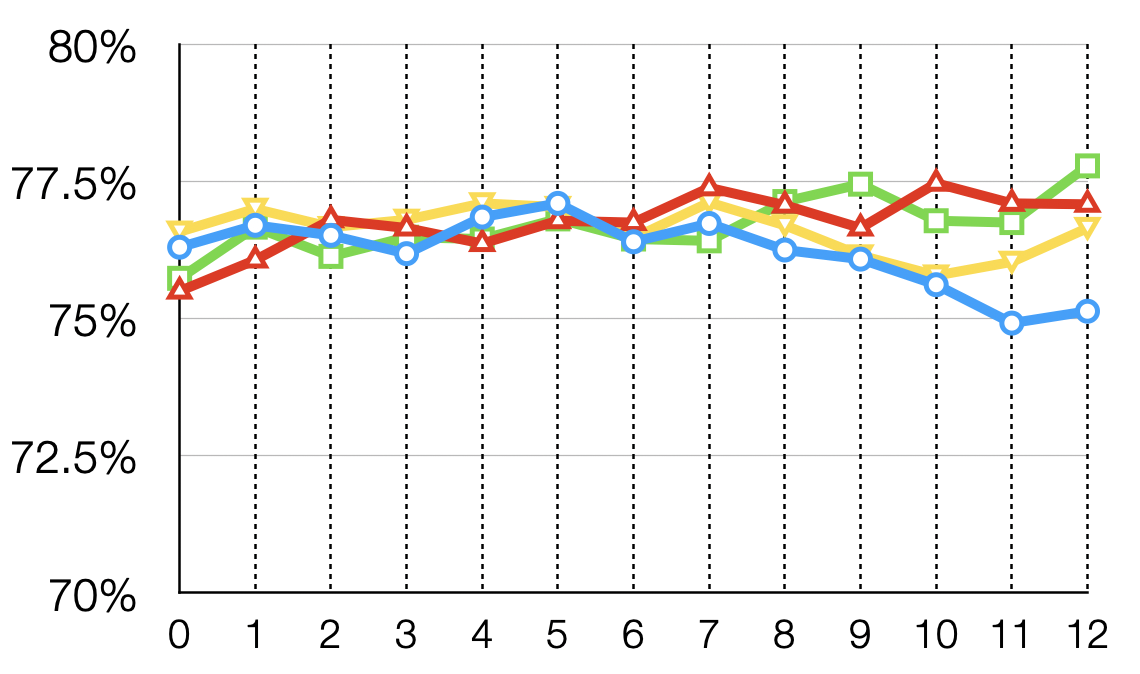}
    \caption{BERT}
    \label{fig:bertavgglue}
    \end{subfigure}
    \begin{subfigure}[b]{0.31\linewidth}
    \centering
    \includegraphics[width=\linewidth]{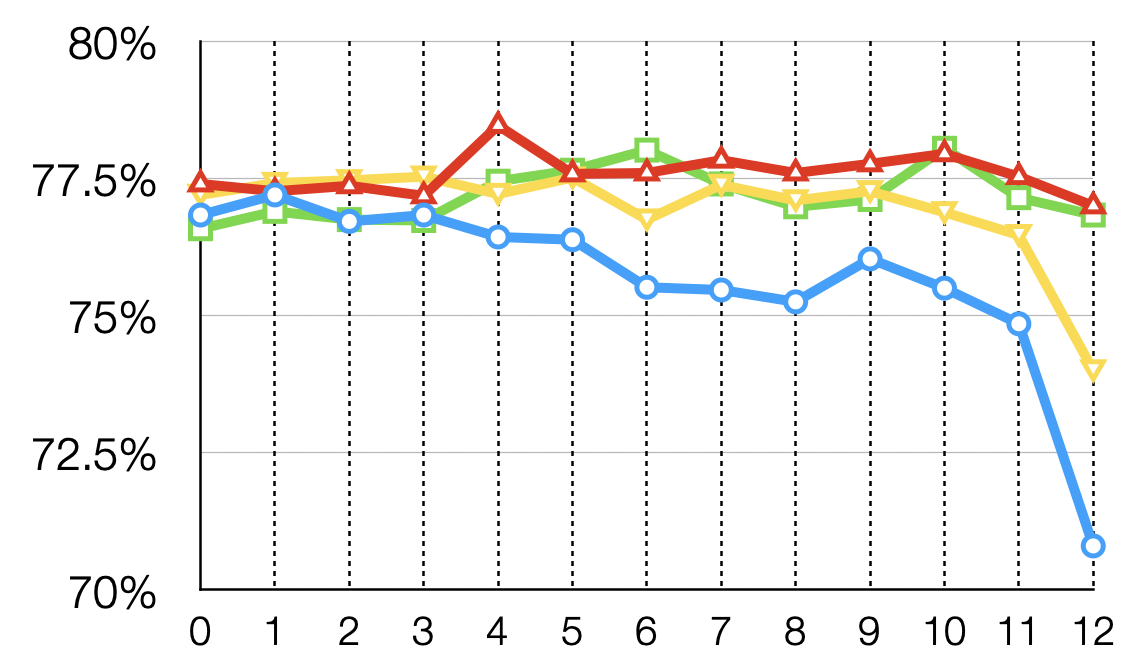}
    \caption{RoBERTa}
    \label{fig:robertaavgglue}
    \end{subfigure}    
    \begin{subfigure}[b]{0.31\linewidth}
    \centering
    \includegraphics[width=\linewidth]{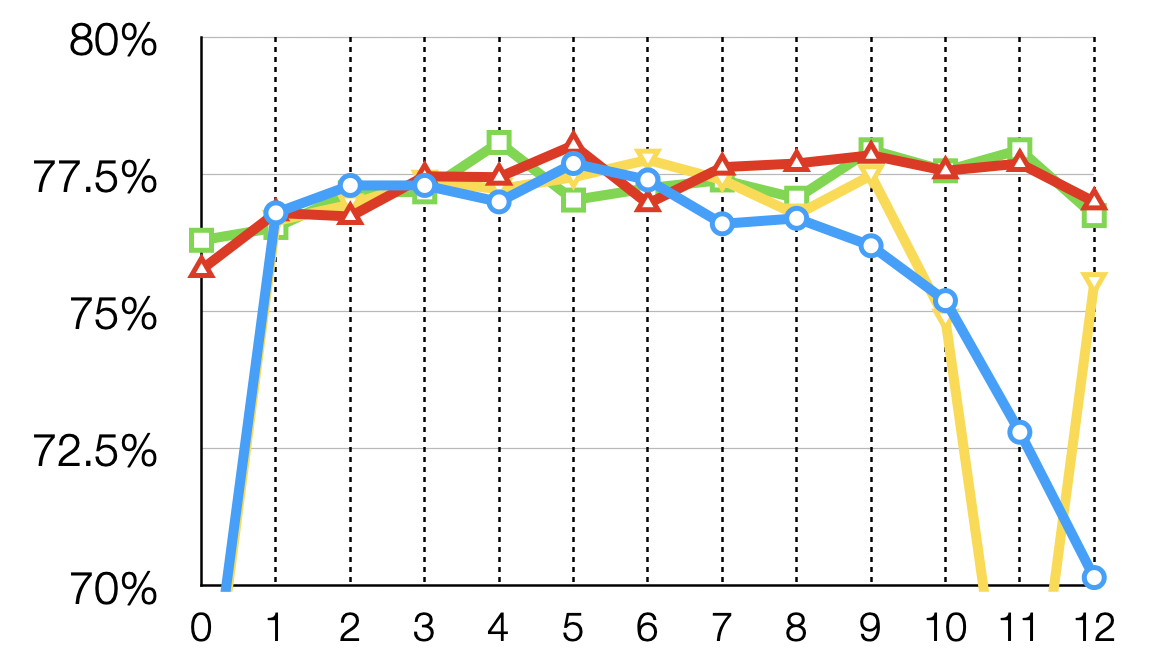}
    \caption{XLNet} 
    \label{fig:xlnetavgglue}
    \end{subfigure}
    \begin{subfigure}[b]{0.45\linewidth}
    \centering
    \includegraphics[width=\linewidth]{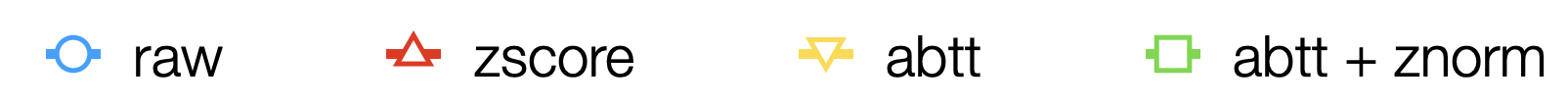}
    \end{subfigure}
    \caption{Average GLUE tasks results}
    \label{fig:avg_glue_results}
\end{figure*}

\subsection{Sequence classification Tasks}
Figure~\ref{fig:avg_glue_results} presents the average layer-wise results using six GLUE tasks.\footnote{We observed similar trends for GPT2 (see Appendix~\ref{sec:app_gpt_glue_tasks}).} The performance improvements when post-processing the higher layers are found to be statistically significant at p=0.05. Additionally, the embedding layer of XLNet and middle to higher layers of RoBERTa achieved statistically significant improvements. Due to the poor performance of \ulen and \minmax, we did not report and discuss their results.

\textbf{Post-processing is generally helpful.} Similar to the performance on the lexical tasks, we observed that \uvar and \abtt post-processing methods resulted in competitive or better performance than \raw. Particularly, \uvar substantially improved the performance of the middle and higher layers (see red line and blue line representing \uvar and \raw respectively). 
\abtt has comparable or better results than \raw, although it never outperformed \uvar. An interesting observation is the embedding layer where \abtt resulted in similar performance to \raw across all models. The embedding layer may encode information related to word identity and position, as in the case of BERT and RoBERTa which is neither useful nor harmful for the downstream tasks. \abtt removed these high principle components while maintaining the performance of the embedding layer. 
%
Combining both post-processings did not result in any consistent benefit over using \uvar alone. 

\textbf{All layers possess information about the tasks.} In contrast to the common notion~\cite{kovaleva-etal-2019-revealing} that the last layer is optimized for the objective function, and hence it is sub-optimal to use for down-stream tasks, we found that after \uvar, the results of the last layers substantially improved, showing competitive results to the best performing layer for each model. 


\textbf{Task-wise performance} Overall, majority of the tasks showed significant improvement with the post-processing of higher-layers. Additionally the embedding layer of XLNet benefited substantially with \znorm. For example, the QNLI performance improved from 69.6 to 80.7 for the embedding layer, and 66.6 to 82.2 for the last layer. The only exception is the SST task that showed minimum benefit of the post-processing methods across all models. The performance differences between \raw and post-processing methods are within 1\% range, and are found to be insignificant.

\begin{figure*}[t]
    \centering
    \begin{subfigure}[b]{0.45\linewidth}
    \centering
    \includegraphics[width=\linewidth]{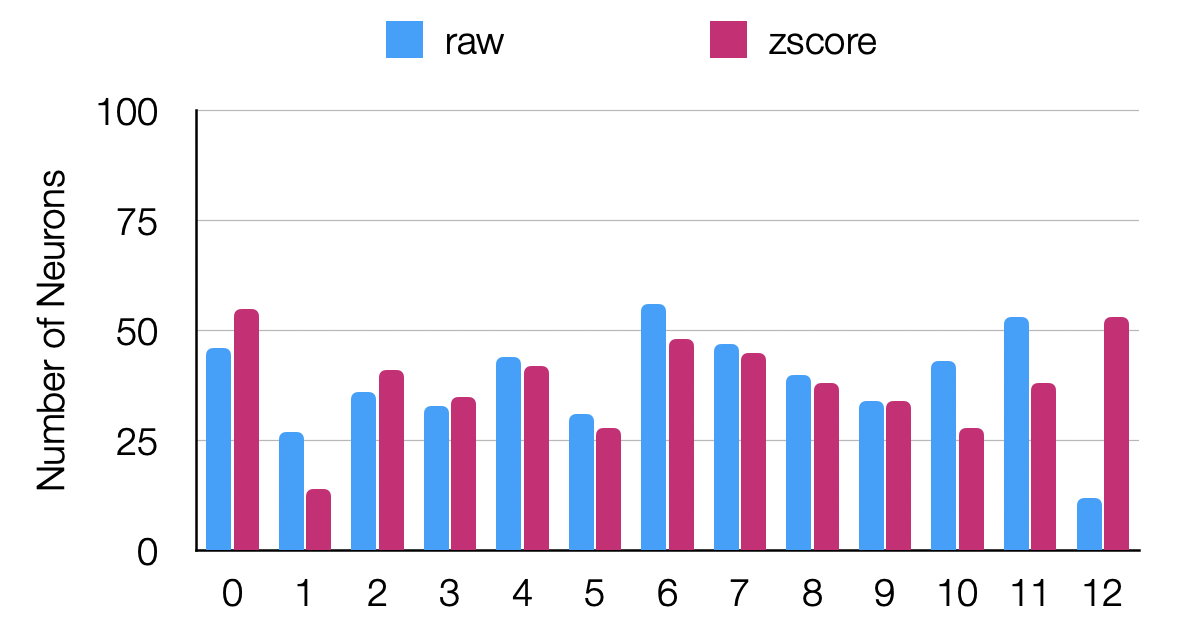}
    \caption{BERT}
    \label{fig:posbert}
    \end{subfigure}
    \hphantom{...}
    \begin{subfigure}[b]{0.43\linewidth}
    \centering
    \includegraphics[width=\linewidth]{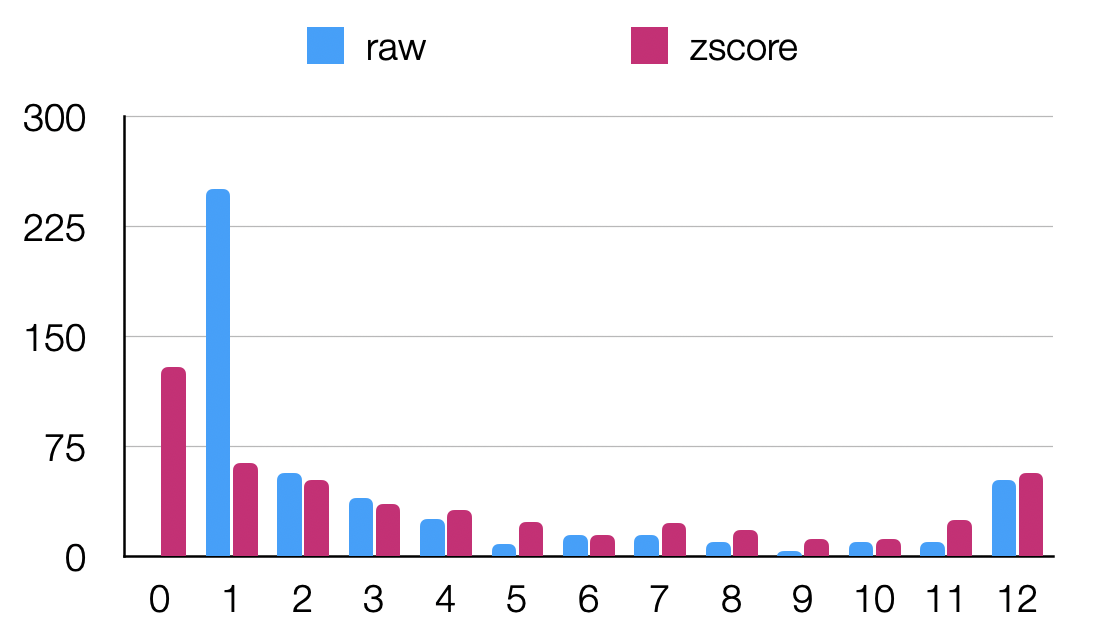}
    \caption{XLNet}
    \label{fig:posxlnet}
    \end{subfigure}
    \caption{Layer-wise contribution of neurons with respect to POS tagging. x-axis corresponds to layers.}
    \label{fig:individual_neurons}
\end{figure*}

\section{Application and Discussion}
\label{sec:discussion}

The effectiveness of the post-processing of representations, particularly \uvar, raises several interesting points in relation to the studies that use contextualized representations like feature-based transfer learning \cite{dalvi-etal-2020-analyzing} and analysis/interpretation of deep models. For example, the work on probing layer representations typically builds a linear classifier and uses the classifier's performance as a proxy to judge how much linguistic information is learned in the representation. Our results show that even when using a strong classification model based on BiLSTM, it is essential to normalize the representations before training. A linear classifier is likely to be more vulnerable to the variance in the features and may not capture the true potential present in the representation. 
Similarly, feature-based transfer learning is directly affected by this post-processing and is likely to improve performance as shown by our sequence classification results. 

In order to probe whether the post-processing of representations would impact representation analysis, we conducted a preliminary experiment on analyzing individual neurons in pre-trained models. \newcite{durrani-2020-individualNeurons} used the Linguistic Correlation Analysis (\texttt{LCA}) method to identify a set of neurons with respect to a linguistic property. The method trains a linear classifier on the linguistic property of interest while using neurons of the pre-trained model as features (12 layers x 768 dimensions = 9984 features). The output of the method results in a list of salient neurons with respect to the property in hand. We consider part-of-speech tagging (POS) as our linguistic property of interest and reproduce their results using the \raw and the \znorm post-processed contextualized embeddings of BERT and XLNet. Since \texttt{LCA} considers contextualized embeddings, we did not aggregate the contextualized embeddings of a word into a single word embedding as describe in Section~\ref{sec:wordrepresentations}. Thus, we use the contextualized embedding extracted from pre-trained models directly in the training of the linear classifier.

Figure \ref{fig:individual_neurons} presents the layer-wise distribution of salient neurons identified by the algorithm using \raw and \znorm contextualized embeddings.
On BERT, the most surprising result is the contribution of last layer which was minimum in the case of \raw but after \znorm, it is among the top contributing layers from which the most salient POS neurons are selected. The results of XLNet are also interesting. The contribution of embedding layer's neuron (0 index on x-axis) is zero in the case of \raw contextualized embeddings while the first layer dominates the distribution. The \znorm contextualized embedding picked the most number of salient neurons from the embedding layer and selected relatively less neurons from the first layer. 
%
This result shows that any analysis obtained by applying external probes on the features generated from pre-trained models needs to consider the effect of normalization into account as it provides an alternative view. In the future, we plan to extend this investigation further by analyzing the effect of post-processing on various other similarity and interpretation analysis works.

While the benefits of post-processing is evident in our experiments, the choice of when to use post-processing is application dependent. In analyzing representations of a network using a classification model, it is recommended to have standardized features to learn the best model. The choice of learned model architecture also plays a role here. The methods invariant to affine transformations would be least effected by the variability of features~\cite{books/wi/KaufmanR90}. For other applications e.g., identifying the importance of a neuron in a pre-trained model, the variance in the values of a neuron can be a signal of its importance and post-processing like \znorm would result in the loss of such information. 


\section{Conclusion}
\label{ssec:conclusion}

We analyzed the effect of four post-processing methods on the contextualized representations using both the lexical and sequence classification tasks. We showed that for lexical tasks, post-processing methods \uvar and \abtt are essential to achieve better performance. On the sequence classification tasks, \uvar alone outperformed all post-processing methods and \raw. The most astonishing results are the large improvements in the performance of the last layers which reflect that 
they
also possess equal amount of information about the lexical and classification tasks but the information is not readily available when used in feature-based transfer learning. 

Our work 
opens several interesting 
frontiers with respect to the work that uses contextualized representations. In a preliminary experiment on representation analysis, we showed that post-processing the representations resulted in different findings. We suggested \uvar as an essential step to consider when using contextualized embeddings for the feature-based transfer learning. 

\vspace{-1mm}
\section*{Ethics and Broader Impact}

We used publicly available datasets, following their terms in the licenses. This includes seven datasets for similarity tasks, three datasets for analogy tasks, and six GLUE datasets. We do not see any harm or ethical issues resulting from our study and findings. Our study benefits the feature-based transfer learning at large and has direct implications towards the work on interpreting and analyzing deep models. 

\bibliography{bib/anthology,bib/custom}
\bibliographystyle{acl_natbib}

\newpage
\clearpage
\section*{Appendix}
\label{sec:appendix}
\appendix
\section{Data Statistics}

In table \ref{tab:sentence_data_statistics}, we present statistics of the different datasets that we used for the experiments. 

\begin{table}[h]                                    
\centering                
\footnotesize
\setlength{\tabcolsep}{2.0pt}
    \begin{tabular}{lrr|lrr}                                    
        \toprule                                    
        Task    & Train & Dev & Task    & Train & Dev  \\        
        \midrule
        SST-2 &  67,349 &   872 & QQP   & 363,846 & 40,430  \\
        MRPC  &   3,668 &   408 & RTE   &   2,490 &   277  \\
        MNLI  & 392,702 &  9,815 & STS-B &   5,749 &  1,500 \\
        QNLI  & 104,743 &  5,463  \\
    \bottomrule
    \end{tabular}
\caption{Data statistics (number of sequences) on the official training and development sets used in the experiments. All are binary classification tasks, except for STS-B, which is a regression task. Recall that the test sets are not publicly available, and hence we use 10\% of the official train as development, and the official development set as our test set. Exact split information is provided in the code README. The data is available at \url{https://gluebenchmark.com/tasks.}
}
\vspace{-6pt}
\label{tab:sentence_data_statistics}                        
\end{table}

\section{GPT-2 Average Results on GLUE}
\label{sec:app_gpt_glue_tasks}
\begin{figure}[h]
    \centering
    \begin{subfigure}{0.4\textwidth} 
    \centering
    \includegraphics[width=\linewidth]{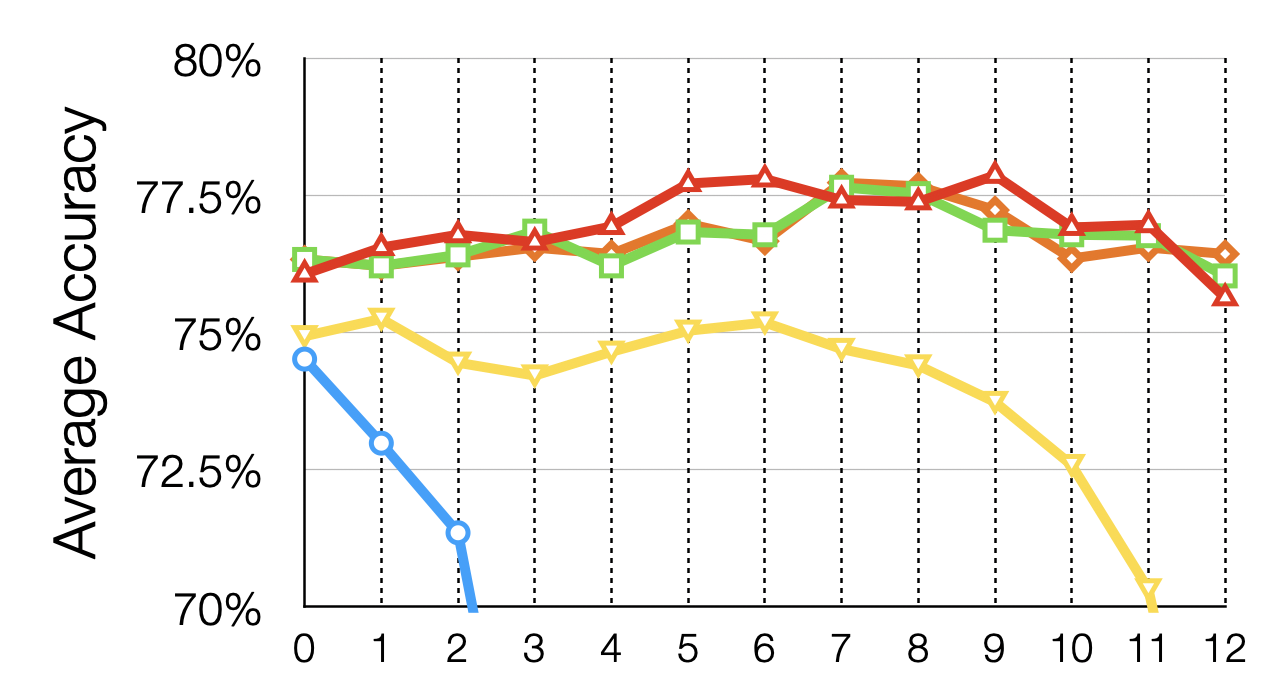}
    \caption{Average GLUE results}
    \label{fig:gptavgglue}
    \end{subfigure}
    
    \begin{subfigure}{0.4\textwidth}
    \centering
    \includegraphics[width=\linewidth]{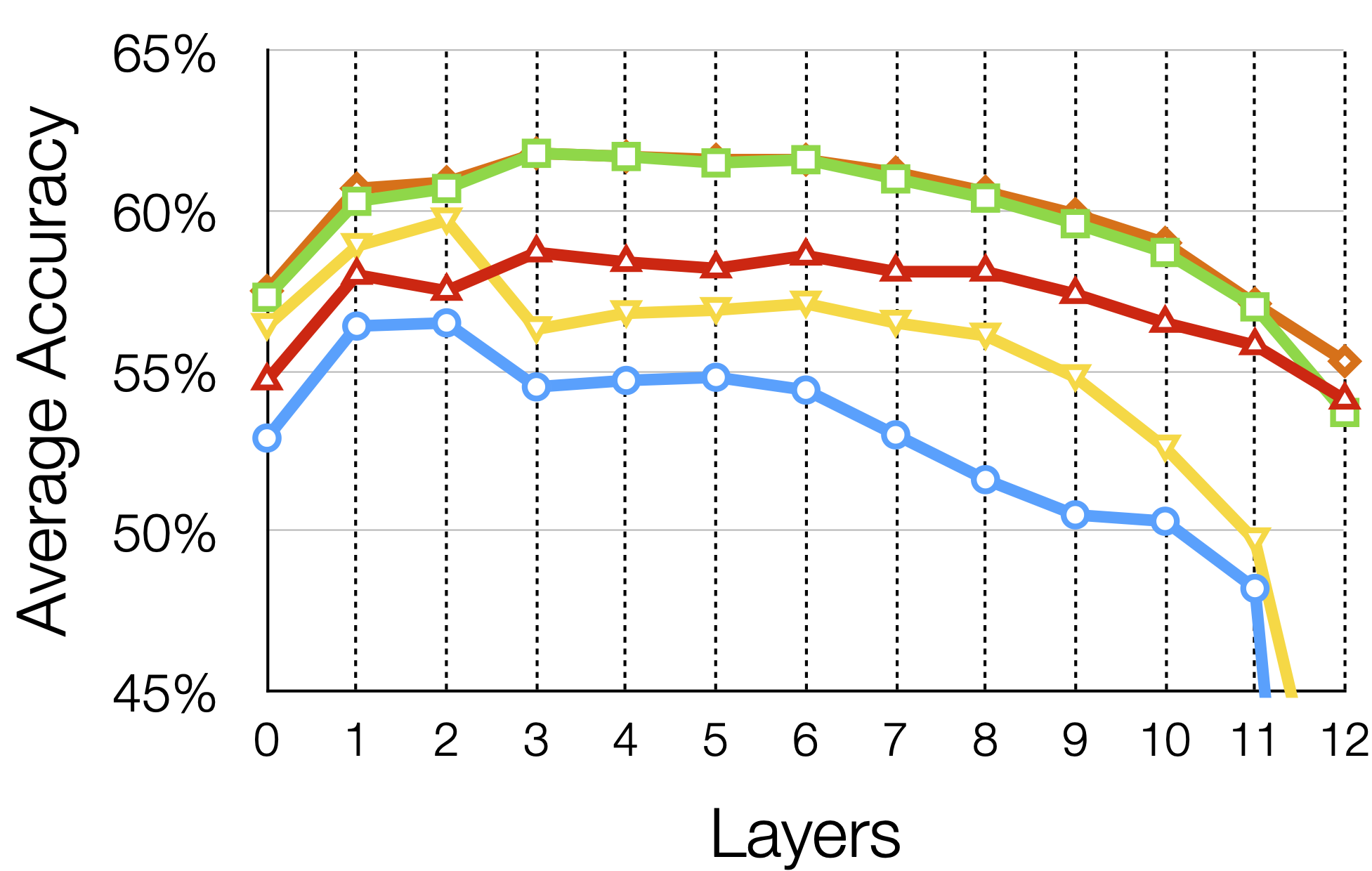}
    \caption{Average Lexical task results}
    \label{fig:gpt_avg_intrinsic}
    \end{subfigure}
    
    \begin{subfigure}{0.4\textwidth} 
    \centering
    \includegraphics[width=\linewidth]{figures/avg_glue_legend.png}
    \end{subfigure}
    \vspace{-5pt}
    \caption{GPT2}
    \label{appendixfig:avg_glue_results}
\end{figure}

In Figure \ref{fig:gptavgglue} and \ref{fig:gpt_avg_intrinsic}, we report the average results for different GLUE and lexical tasks across different L/Ns, respectively.

\section{Sequence Classification Results}
\label{sec:app_seq_classification_results}
In the following tables, we provide results obtained from raw embedding, z-socre normalization and all-but-the-top, for \textbf{BERT:} \ref{table:BERT_raw_embedding}, \ref{table:BERT_z-score_normalization}, \ref{table:BERT_all-but-the-top}; \textbf{XLNet:} \ref{table:XLNet_raw_embedding}, \ref{table:XLNet_z-score_normalization}, \ref{table:XLNet_all-but-the-top}; \textbf{RoBERTa:} \ref{table:RoBERTa_z-score_normalization}, \ref{table:RoBERTa_all-but-the-top} and \textbf{GPT2:} \ref{table:GPT2_raw embedding}, \ref{table:GPT2_z-score_normalization}, \ref{table:GPT2_all-but-the-top}.

\begin{table}[tbh!]
\centering
\footnotesize
\setlength{\tabcolsep}{3pt}
\scalebox{0.75}{
\begin{tabular}{llllllll}
\toprule
\textbf{L/N} & \textbf{QNLI}  & \textbf{MNLI}  & \textbf{MRPC}  & \textbf{SST}   & \textbf{STS}   & \textbf{RTE}   & \textbf{RTE}   \\ \midrule
0      & 0.82  & 0.724 & 0.775 & 0.88  & 0.801 & 0.578 & 0.578 \\
1      & 0.815 & 0.724 & 0.797 & 0.885 & 0.8   & 0.581 & 0.581 \\
2      & 0.816 & 0.712 & 0.784 & 0.89  & 0.808 & 0.581 & 0.581 \\
3      & 0.805 & 0.716 & 0.779 & 0.894 & 0.807 & 0.57  & 0.57  \\
4      & 0.818 & 0.734 & 0.775 & 0.89  & 0.806 & 0.588 & 0.588 \\
5      & 0.824 & 0.74  & 0.779 & 0.901 & 0.804 & 0.578 & 0.578 \\
6      & 0.813 & 0.739 & 0.777 & 0.896 & 0.799 & 0.56  & 0.56  \\
7      & 0.821 & 0.735 & 0.777 & 0.888 & 0.802 & 0.581 & 0.581 \\
8      & 0.813 & 0.733 & 0.784 & 0.882 & 0.8   & 0.563 & 0.563 \\
9      & 0.8   & 0.726 & 0.784 & 0.896 & 0.799 & 0.56  & 0.56  \\
10     & 0.794 & 0.725 & 0.787 & 0.889 & 0.8   & 0.542 & 0.542 \\
11     & 0.773 & 0.716 & 0.772 & 0.898 & 0.794 & 0.542 & 0.542 \\
12     & 0.805 & 0.728 & 0.77  & 0.878 & 0.785 & 0.542 & 0.542 \\ \bottomrule
\end{tabular}
}
\caption{\textbf{BERT}: raw embedding}
\label{table:BERT_raw_embedding}
\end{table}

\begin{table}[tbh!]
\centering
\footnotesize
\setlength{\tabcolsep}{3pt}
\scalebox{0.8}{
\begin{tabular}{llllllll}
\toprule
\textbf{L/N} & \textbf{QNLI}  & \textbf{MNLI}  & \textbf{MRPC}  & \textbf{SST}   & \textbf{STS corr} & \textbf{RTE}   & \textbf{uvar}  \\ \midrule
0      & 0.801 & 0.705 & 0.77  & 0.881 & 0.792    & 0.581 & 0.755 \\
1      & 0.802 & 0.727 & 0.765 & 0.885 & 0.797    & 0.588 & 0.761 \\
2      & 0.818 & 0.728 & 0.784 & 0.886 & 0.804    & 0.588 & 0.768 \\
3      & 0.809 & 0.71  & 0.775 & 0.885 & 0.81     & 0.61  & 0.767 \\
4      & 0.807 & 0.727 & 0.762 & 0.886 & 0.804    & 0.596 & 0.764 \\
5      & 0.813 & 0.729 & 0.772 & 0.896 & 0.812    & 0.585 & 0.768 \\
6      & 0.816 & 0.726 & 0.775 & 0.901 & 0.813    & 0.574 & 0.768 \\
7      & 0.822 & 0.726 & 0.799 & 0.892 & 0.816    & 0.588 & 0.774 \\
8      & 0.813 & 0.742 & 0.787 & 0.892 & 0.816    & 0.574 & 0.771 \\
9      & 0.811 & 0.734 & 0.77  & 0.897 & 0.82     & 0.567 & 0.767 \\
10     & 0.812 & 0.734 & 0.797 & 0.894 & 0.819    & 0.592 & 0.775 \\
11     & 0.808 & 0.733 & 0.789 & 0.901 & 0.817    & 0.578 & 0.771 \\
12     & 0.808 & 0.733 & 0.799 & 0.883 & 0.821    & 0.581 & 0.771 \\ \bottomrule
\end{tabular}
}
\caption{\textbf{BERT}: z-score normalization}
\label{table:BERT_z-score_normalization}
\end{table}

\begin{table}[tbh!]
\centering
\footnotesize
\setlength{\tabcolsep}{3pt}
\scalebox{0.8}{
\begin{tabular}{llllllll}
\toprule
\textbf{L/N} & \textbf{QNLI}  & \textbf{MNLI}  & \textbf{MRPC}  & \textbf{SST}   & \textbf{STS corr} & \textbf{RTE}   & \textbf{abtt}  \\ \midrule
0                    & 0.82  & 0.72  & 0.772 & 0.872 & 0.801    & 0.61  & 0.766 \\
1                    & 0.823 & 0.729 & 0.779 & 0.882 & 0.811    & 0.596 & 0.770 \\
2                    & 0.808 & 0.727 & 0.777 & 0.881 & 0.811    & 0.596 & 0.767 \\
3                    & 0.817 & 0.732 & 0.784 & 0.886 & 0.815    & 0.574 & 0.768 \\
4                    & 0.819 & 0.724 & 0.789 & 0.885 & 0.81     & 0.599 & 0.771 \\
5                    & 0.824 & 0.73  & 0.775 & 0.89  & 0.807    & 0.596 & 0.770 \\
6                    & 0.822 & 0.73  & 0.775 & 0.888 & 0.81     & 0.563 & 0.765 \\
7                    & 0.829 & 0.729 & 0.784 & 0.886 & 0.814    & 0.585 & 0.771 \\
8                    & 0.817 & 0.734 & 0.775 & 0.896 & 0.82     & 0.56  & 0.767 \\
9                    & 0.813 & 0.737 & 0.775 & 0.891 & 0.811    & 0.542 & 0.762 \\
10                   & 0.815 & 0.732 & 0.772 & 0.896 & 0.801    & 0.531 & 0.758 \\
11                   & 0.814 & 0.726 & 0.779 & 0.894 & 0.807    & 0.542 & 0.760 \\
12                   & 0.824 & 0.746 & 0.782 & 0.885 & 0.813    & 0.549 & 0.767 \\ \bottomrule
\end{tabular}
}
\caption{\textbf{BERT}: all-but-the-top}
\label{table:BERT_all-but-the-top}
\vspace{-6pt}
\end{table}

\begin{table}[tbh!]
\centering
\footnotesize
\setlength{\tabcolsep}{3pt}
\scalebox{0.85}{
\begin{tabular}{lllllll}
\toprule
\textbf{L/N} & \textbf{QNLI}  & \textbf{MNLI}  & \textbf{MRPC}  & \textbf{SST}   & \textbf{STS corr} & \textbf{RTE}   \\ \midrule
0      & 0.696 & 0.705 & 0.691 & 0.853 & 0.52     & 0.531 \\
1      & 0.81  & 0.717 & 0.789 & 0.892 & 0.812    & 0.588 \\
2      & 0.81  & 0.735 & 0.797 & 0.891 & 0.814    & 0.592 \\
3      & 0.803 & 0.736 & 0.777 & 0.898 & 0.82     & 0.603 \\
4      & 0.791 & 0.735 & 0.777 & 0.893 & 0.824    & 0.599 \\
5      & 0.803 & 0.729 & 0.801 & 0.888 & 0.818    & 0.625 \\
6      & 0.813 & 0.734 & 0.787 & 0.901 & 0.805    & 0.606 \\
7      & 0.812 & 0.733 & 0.784 & 0.901 & 0.807    & 0.556 \\
8      & 0.782 & 0.726 & 0.814 & 0.896 & 0.79     & 0.596 \\
9      & 0.794 & 0.723 & 0.797 & 0.893 & 0.803    & 0.563 \\
10     & 0.76  & 0.724 & 0.772 & 0.897 & 0.791    & 0.57  \\
11     & 0.722 & 0.669 & 0.777 & 0.884 & 0.75     & 0.563 \\
12     & 0.666 & 0.641 & 0.733 & 0.892 & 0.732    & 0.545 \\ \bottomrule
\end{tabular}
}
\caption{\textbf{XLNet}: raw embedding}
\label{table:XLNet_raw_embedding}
\end{table}

\begin{table}[tbh!]
\centering
\footnotesize
\setlength{\tabcolsep}{3pt}
\scalebox{0.9}{
\begin{tabular}{lllllll}
\toprule
\textbf{L/N} & \textbf{QNLI}  & \textbf{MNLI}  & \textbf{MRPC}  & \textbf{SST}   & \textbf{STS corr} & \textbf{RTE}   \\ \midrule
0      & 0.807 & 0.722 & 0.775 & 0.885 & 0.805    & 0.552 \\
1      & 0.806 & 0.727 & 0.792 & 0.896 & 0.809    & 0.578 \\
2      & 0.807 & 0.733 & 0.777 & 0.897 & 0.816    & 0.574 \\
3      & 0.815 & 0.742 & 0.772 & 0.898 & 0.822    & 0.599 \\
4      & 0.816 & 0.736 & 0.77  & 0.894 & 0.828    & 0.603 \\
5      & 0.821 & 0.733 & 0.792 & 0.896 & 0.826    & 0.614 \\
6      & 0.813 & 0.715 & 0.782 & 0.896 & 0.827    & 0.585 \\
7      & 0.815 & 0.738 & 0.787 & 0.894 & 0.825    & 0.599 \\
8      & 0.815 & 0.735 & 0.797 & 0.9   & 0.823    & 0.592 \\
9      & 0.808 & 0.735 & 0.799 & 0.9   & 0.823    & 0.606 \\
10     & 0.809 & 0.733 & 0.806 & 0.89  & 0.82     & 0.596 \\
11     & 0.821 & 0.746 & 0.799 & 0.89  & 0.828    & 0.578 \\
12     & 0.822 & 0.74  & 0.792 & 0.886 & 0.824    & 0.556 \\ \bottomrule
\end{tabular}
}
\caption{\textbf{XLNet}: z-score normalization}
\label{table:XLNet_z-score_normalization}
\end{table}

\begin{table}[tbh!]
\centering
\footnotesize
\setlength{\tabcolsep}{3pt}
\scalebox{0.9}{
\begin{tabular}{lllllll}
\toprule
\textbf{L/N} & \textbf{QNLI}  & \textbf{MNLI}  & \textbf{MRPC}  & \textbf{SST}   & \textbf{STS corr} & \textbf{RTE}   \\ \midrule
0      & 0.661 & 0.708 & 0.703 & 0.853 & 0.509    & 0.542 \\
1      & 0.803 & 0.724 & 0.77  & 0.891 & 0.815    & 0.596 \\
2      & 0.823 & 0.735 & 0.76  & 0.899 & 0.819    & 0.581 \\
3      & 0.823 & 0.731 & 0.794 & 0.89  & 0.824    & 0.581 \\
4      & 0.813 & 0.744 & 0.794 & 0.888 & 0.825    & 0.57  \\
5      & 0.824 & 0.736 & 0.792 & 0.897 & 0.824    & 0.574 \\
6      & 0.824 & 0.733 & 0.789 & 0.892 & 0.829    & 0.599 \\
7      & 0.827 & 0.732 & 0.801 & 0.891 & 0.826    & 0.567 \\
8      & 0.817 & 0.729 & 0.799 & 0.893 & 0.812    & 0.556 \\
9      & 0.824 & 0.726 & 0.809 & 0.892 & 0.813    & 0.585 \\
10     & 0.789 & 0.693 & 0.779 & 0.891 & 0.812    & 0.527 \\
11     & 0.657 & 0.6   & 0.721 & 0.876 & 0.536    & 0.527 \\
12     & 0.805 & 0.696 & 0.77  & 0.89  & 0.828    & 0.542 \\ \bottomrule
\end{tabular}
}
\caption{\textbf{XLNet}: all-but-the-top}
\label{table:XLNet_all-but-the-top}
\end{table}

\begin{table}[tbh!]
\centering
\footnotesize
\setlength{\tabcolsep}{3pt}
\scalebox{0.87}{
\begin{tabular}{lllllll}
\toprule
\textbf{L/N} & \textbf{QNLI}  & \textbf{MNLI}  & \textbf{MRPC}  & \textbf{SST}   & \textbf{STS corr} & \textbf{RTE}   \\ \midrule
0      & 0.817 & 0.742 & 0.784 & 0.875 & 0.818    & 0.574 \\
1      & 0.814 & 0.724 & 0.806 & 0.891 & 0.809    & 0.588 \\
2      & 0.812 & 0.726 & 0.789 & 0.893 & 0.809    & 0.574 \\
3      & 0.813 & 0.74  & 0.787 & 0.882 & 0.803    & 0.585 \\
4      & 0.807 & 0.722 & 0.775 & 0.884 & 0.806    & 0.592 \\
5      & 0.804 & 0.722 & 0.775 & 0.896 & 0.798    & 0.588 \\
6      & 0.794 & 0.725 & 0.775 & 0.892 & 0.785    & 0.56  \\
7      & 0.795 & 0.72  & 0.782 & 0.883 & 0.788    & 0.56  \\
8      & 0.792 & 0.726 & 0.775 & 0.878 & 0.784    & 0.56  \\
9      & 0.789 & 0.732 & 0.775 & 0.883 & 0.787    & 0.596 \\
10     & 0.793 & 0.716 & 0.792 & 0.893 & 0.791    & 0.545 \\
11     & 0.783 & 0.715 & 0.782 & 0.892 & 0.785    & 0.534 \\
12     & 0.732 & 0.708 & 0.755 & 0.865 & 0.643    & 0.545 \\ \bottomrule
\end{tabular}
}
\caption{\textbf{RoBERTa}: raw embedding}
\label{table:RoBERTa_raw_embedding}
\end{table}

\begin{table}[tbh!]
\centering
\footnotesize
\setlength{\tabcolsep}{3pt}
\scalebox{0.87}{
\begin{tabular}{lllllll}
\toprule
\textbf{L/N} & \textbf{QNLI}  & \textbf{MNLI}  & \textbf{MRPC}  & \textbf{SST}   & \textbf{STS corr} & \textbf{RTE}   \\ \midrule
0      & 0.824 & 0.735 & 0.789 & 0.89  & 0.803    & 0.603 \\
1      & 0.815 & 0.729 & 0.792 & 0.903 & 0.816    & 0.581 \\
2      & 0.819 & 0.729 & 0.787 & 0.901 & 0.814    & 0.592 \\
3      & 0.824 & 0.73  & 0.772 & 0.901 & 0.823    & 0.581 \\
4      & 0.825 & 0.728 & 0.804 & 0.905 & 0.829    & 0.617 \\
5      & 0.817 & 0.734 & 0.792 & 0.9   & 0.824    & 0.588 \\
6      & 0.815 & 0.73  & 0.799 & 0.894 & 0.826    & 0.592 \\
7      & 0.818 & 0.732 & 0.804 & 0.9   & 0.824    & 0.592 \\
8      & 0.815 & 0.73  & 0.804 & 0.884 & 0.824    & 0.599 \\
9      & 0.821 & 0.742 & 0.792 & 0.901 & 0.825    & 0.585 \\
10     & 0.81  & 0.739 & 0.794 & 0.897 & 0.823    & 0.614 \\
11     & 0.817 & 0.73  & 0.797 & 0.894 & 0.826    & 0.588 \\
12     & 0.814 & 0.737 & 0.792 & 0.89  & 0.827    & 0.56  \\ \bottomrule
\end{tabular}
}
\caption{\textbf{RoBERTa}: z-score normalization}
\label{table:RoBERTa_z-score_normalization}
\end{table}

\begin{table}[tbh!]
\centering
\footnotesize
\setlength{\tabcolsep}{3pt}
\scalebox{0.75}{
\begin{tabular}{lllllll}
\toprule
\textbf{L/N} & \textbf{QNLI}  & \textbf{MNLI}  & \textbf{MRPC}  & \textbf{SST}   & \textbf{STS corr} & \textbf{RTE}   \\ \midrule
0      & 0.814 & 0.734 & 0.797 & 0.876 & 0.826    & 0.585 \\
1      & 0.821 & 0.744 & 0.809 & 0.891 & 0.813    & 0.567 \\
2      & 0.819 & 0.727 & 0.799 & 0.908 & 0.81     & 0.585 \\
3      & 0.826 & 0.738 & 0.797 & 0.889 & 0.817    & 0.585 \\
4      & 0.823 & 0.737 & 0.787 & 0.888 & 0.817    & 0.581 \\
5      & 0.824 & 0.736 & 0.806 & 0.89  & 0.821    & 0.574 \\
6      & 0.813 & 0.733 & 0.792 & 0.896 & 0.808    & 0.563 \\
7      & 0.818 & 0.739 & 0.797 & 0.885 & 0.812    & 0.592 \\
8      & 0.817 & 0.738 & 0.794 & 0.884 & 0.815    & 0.578 \\
9      & 0.816 & 0.741 & 0.789 & 0.884 & 0.814    & 0.592 \\
10     & 0.817 & 0.74  & 0.787 & 0.896 & 0.799    & 0.574 \\
11     & 0.813 & 0.733 & 0.789 & 0.881 & 0.805    & 0.567 \\
12     & 0.794 & 0.716 & 0.767 & 0.876 & 0.745    & 0.542 \\ \bottomrule
\end{tabular}
}
\caption{\textbf{RoBERTa}: all-but-the-top}
\label{table:RoBERTa_all-but-the-top}
\end{table}

\begin{table}[tbh!]
\centering
\footnotesize
\scalebox{0.75}{
\begin{tabular}{lllllll}
\toprule
\textbf{L/N} & \textbf{QNLI}  & \textbf{MNLI}  & \textbf{MRPC}  & \textbf{SST}   & \textbf{STS corr} & \textbf{RTE}   \\ \midrule
0      & 0.804 & 0.719 & 0.75  & 0.868 & 0.763    & 0.567 \\
1      & 0.705 & 0.706 & 0.748 & 0.885 & 0.79     & 0.545 \\
2      & 0.707 & 0.695 & 0.713 & 0.882 & 0.714    & 0.57  \\
3      & 0.674 & 0.663 & 0.696 & 0.892 & 0.388    & 0.534 \\
4      & 0.667 & 0.633 & 0.672 & 0.884 & 0.429    & 0.556 \\
5      & 0.667 & 0.648 & 0.699 & 0.901 & 0.414    & 0.563 \\
6      & 0.65  & 0.655 & 0.689 & 0.901 & 0.428    & 0.542 \\
7      & 0.662 & 0.649 & 0.708 & 0.898 & 0.252    & 0.57  \\
8      & 0.655 & 0.665 & 0.672 & 0.896 & 0.376    & 0.581 \\
9      & 0.658 & 0.584 & 0.686 & 0.894 & 0.268    & 0.567 \\
10     & 0.665 & 0.663 & 0.676 & 0.893 & 0.313    & 0.538 \\
11     & 0.666 & 0.647 & 0.689 & 0.892 & 0.353    & 0.531 \\ \bottomrule
12     & 0.638 & 0.623 & 0.703 & 0.847 & 0.274    & 0.538
\end{tabular}
}
\caption{\textbf{GPT-2}: raw embedding}
\label{table:GPT2_raw embedding}
\end{table}

\begin{table}[tbh!]
\centering
\footnotesize
\setlength{\tabcolsep}{3pt}
\scalebox{0.85}{
\begin{tabular}{lllllll}
\toprule
\textbf{L/N} & \textbf{QNLI}  & \textbf{MNLI}  & \textbf{MRPC}  & \textbf{SST}   & \textbf{STS corr} & \textbf{RTE}   \\ \midrule
0      & 0.805 & 0.724 & 0.784 & 0.891 & 0.811    & 0.549 \\
1      & 0.801 & 0.738 & 0.777 & 0.89  & 0.817    & 0.57  \\
2      & 0.805 & 0.742 & 0.789 & 0.881 & 0.82     & 0.57  \\
3      & 0.803 & 0.734 & 0.789 & 0.888 & 0.818    & 0.567 \\
4      & 0.809 & 0.722 & 0.804 & 0.885 & 0.818    & 0.578 \\
5      & 0.817 & 0.743 & 0.792 & 0.898 & 0.821    & 0.592 \\
6      & 0.816 & 0.73  & 0.804 & 0.898 & 0.824    & 0.596 \\
7      & 0.81  & 0.738 & 0.787 & 0.892 & 0.822    & 0.596 \\
8      & 0.81  & 0.72  & 0.806 & 0.889 & 0.822    & 0.596 \\
9      & 0.812 & 0.736 & 0.784 & 0.9   & 0.819    & 0.621 \\
10     & 0.804 & 0.728 & 0.794 & 0.888 & 0.813    & 0.588 \\
11     & 0.798 & 0.726 & 0.787 & 0.892 & 0.809    & 0.606 \\
12     & 0.796 & 0.717 & 0.779 & 0.881 & 0.802    & 0.563 \\ \bottomrule
\end{tabular}
}
\caption{\textbf{GPT-2}: z-score normalization}
\label{table:GPT2_z-score_normalization}

\end{table}

\begin{table}[tbh!]
\centering
\footnotesize
\setlength{\tabcolsep}{3pt}
\scalebox{0.85}{
\begin{tabular}{lllllll}
\toprule
\textbf{L/N} & \textbf{QNLI}  & \textbf{MNLI}  & \textbf{MRPC}  & \textbf{SST}   & \textbf{STS corr} & \textbf{RTE}   \\ \midrule
0      & 0.803 & 0.732 & 0.765 & 0.865 & 0.786    & 0.545 \\
1      & 0.752 & 0.71  & 0.787 & 0.89  & 0.802    & 0.574 \\
2      & 0.722 & 0.685 & 0.784 & 0.897 & 0.812    & 0.567 \\
3      & 0.757 & 0.708 & 0.76  & 0.881 & 0.795    & 0.552 \\
4      & 0.754 & 0.712 & 0.765 & 0.888 & 0.804    & 0.556 \\
5      & 0.763 & 0.713 & 0.767 & 0.892 & 0.8      & 0.567 \\
6      & 0.77  & 0.717 & 0.762 & 0.89  & 0.812    & 0.56  \\
7      & 0.754 & 0.716 & 0.762 & 0.888 & 0.806    & 0.556 \\
8      & 0.756 & 0.714 & 0.752 & 0.891 & 0.791    & 0.56  \\
9      & 0.739 & 0.697 & 0.765 & 0.893 & 0.774    & 0.556 \\
10     & 0.726 & 0.683 & 0.748 & 0.884 & 0.769    & 0.545 \\
11     & 0.722 & 0.67  & 0.708 & 0.893 & 0.688    & 0.538 \\
12     & 0.677 & 0.599 & 0.713 & 0.85  & 0.47     & 0.531 \\ \bottomrule
\end{tabular}
}
\caption{\textbf{GPT-2}: all-but-the-top}
\label{table:GPT2_all-but-the-top}
\end{table}

\section{Lexical task-wise Results}
\label{sec:app_results_intrinsic_tasks}
In the following tables we provide task-wise results across different results for different models: \textbf{BERT:} \ref{tab:intrinsic_task_bert_raw}, \ref{tab:intrinsic_task_bert_unit_var}, \ref{tab:intrinsic_task_bert_abtt_raw}; \textbf{XLNet:} \ref{tab:intrinsic_task_xlnet_raw}, \ref{tab:intrinsic_task_xlnet_unit_var}, \ref{tab:intrinsic_task_xlnet_abtt}. 

\begin{table*}[tbh!]
\centering
\setlength{\tabcolsep}{3pt}
\scalebox{0.65}{
\begin{tabular}{lrrrrrrrrrrrr}
\toprule
L/N &
  \multicolumn{1}{l}{MEN} &
  \multicolumn{1}{l}{WS353} &
  \multicolumn{1}{l}{WS353R} &
  \multicolumn{1}{l}{WS353S} &
  \multicolumn{1}{l}{SimLex999} &
  \multicolumn{1}{l}{RW} &
  \multicolumn{1}{l}{RG65} &
  \multicolumn{1}{l}{MTurk} &
  \multicolumn{1}{l}{Google} &
  \multicolumn{1}{l}{MSR} &
  \multicolumn{1}{l}{SemEval2012\_2} &
  \multicolumn{1}{l}{Average} \\ \midrule
L0      & 0.617 & 0.543 & 0.606 & 0.326 & 0.445 & 0.539 & 0.515 & 0.609 & 0.324 & 0.706 & 0.220 & 0.495 \\
L1      & 0.671 & 0.530 & 0.635 & 0.369 & 0.462 & 0.600 & 0.575 & 0.671 & 0.341 & 0.729 & 0.222 & 0.528 \\
L2      & 0.683 & 0.514 & 0.639 & 0.390 & 0.481 & 0.617 & 0.590 & 0.689 & 0.364 & 0.749 & 0.222 & 0.540 \\
L3      & 0.678 & 0.508 & 0.672 & 0.406 & 0.488 & 0.624 & 0.593 & 0.710 & 0.369 & 0.733 & 0.224 & 0.546 \\
L4      & 0.679 & 0.482 & 0.719 & 0.419 & 0.499 & 0.620 & 0.569 & 0.734 & 0.379 & 0.721 & 0.233 & 0.550 \\
L5      & 0.677 & 0.463 & 0.768 & 0.441 & 0.502 & 0.612 & 0.535 & 0.744 & 0.395 & 0.748 & 0.233 & 0.556 \\
L6      & 0.666 & 0.454 & 0.777 & 0.440 & 0.508 & 0.587 & 0.494 & 0.723 & 0.395 & 0.754 & 0.242 & 0.549 \\
L7      & 0.687 & 0.472 & 0.789 & 0.449 & 0.524 & 0.608 & 0.514 & 0.741 & 0.397 & 0.765 & 0.239 & 0.562 \\
L8      & 0.713 & 0.498 & 0.822 & 0.464 & 0.539 & 0.636 & 0.558 & 0.751 & 0.397 & 0.767 & 0.229 & 0.579 \\
L9      & 0.710 & 0.478 & 0.844 & 0.457 & 0.537 & 0.613 & 0.543 & 0.708 & 0.399 & 0.774 & 0.220 & 0.571 \\
L10     & 0.686 & 0.444 & 0.825 & 0.444 & 0.500 & 0.579 & 0.513 & 0.666 & 0.398 & 0.772 & 0.221 & 0.550 \\
L11     & 0.654 & 0.406 & 0.824 & 0.414 & 0.479 & 0.528 & 0.477 & 0.605 & 0.402 & 0.781 & 0.215 & 0.526 \\
L12     & 0.643 & 0.406 & 0.771 & 0.413 & 0.488 & 0.558 & 0.517 & 0.620 & 0.406 & 0.820 & 0.188 & 0.530 \\ \midrule
\textbf{Average} & \textbf{0.674} & \textbf{0.477} & \textbf{0.746} & \textbf{0.418} & \textbf{0.496} & \textbf{0.594} & \textbf{0.538} & \textbf{0.690} & \textbf{0.382} & \textbf{0.755} & \textbf{0.224} & \textbf{0.545} \\\midrule
\textbf{Max}     & \textbf{0.713} & \textbf{0.543} & \textbf{0.844} & \textbf{0.464} & \textbf{0.539} & \textbf{0.636} & \textbf{0.593} & \textbf{0.751} & \textbf{0.406} & \textbf{0.820} & \textbf{0.242} & \textbf{0.579} \\\bottomrule
\end{tabular}
}
\caption{\textbf{BERT}: raw embedding}
\label{tab:intrinsic_task_bert_raw}
\end{table*}

\begin{table*}[tbh!]
\centering
\setlength{\tabcolsep}{3pt}
\scalebox{0.65}{
\begin{tabular}{@{}lrrrrrrrrrrrr@{}}
\toprule
\textbf{L/N} &
  \multicolumn{1}{l}{\textbf{MEN}} &
  \multicolumn{1}{l}{\textbf{WS353}} &
  \multicolumn{1}{l}{\textbf{WS353R}} &
  \multicolumn{1}{l}{\textbf{WS353S}} &
  \multicolumn{1}{l}{\textbf{SimLex999}} &
  \multicolumn{1}{l}{\textbf{RW}} &
  \multicolumn{1}{l}{\textbf{RG65}} &
  \multicolumn{1}{l}{\textbf{MTurk}} &
  \multicolumn{1}{l}{\textbf{Google}} &
  \multicolumn{1}{l}{\textbf{MSR}} &
  \multicolumn{1}{l}{\textbf{SemEval2012\_2}} &
  \multicolumn{1}{l}{\textbf{Average}} \\ \midrule
L0 & 0.588 & 0.513 & 0.606 & 0.263 & 0.432 & 0.511 & 0.505 & 0.562 & 0.326 & 0.706 & 0.215 & 0.480 \\
L1 & 0.632 & 0.529 & 0.619 & 0.302 & 0.444 & 0.558 & 0.552 & 0.610 & 0.342 & 0.729 & 0.222 & 0.495 \\
L2 & 0.655 & 0.540 & 0.620 & 0.347 & 0.459 & 0.586 & 0.564 & 0.644 & 0.365 & 0.751 & 0.223 & 0.519 \\
L3 & 0.656 & 0.536 & 0.662 & 0.376 & 0.467 & 0.593 & 0.560 & 0.663 & 0.368 & 0.731 & 0.218 & 0.522 \\
L4 & 0.670 & 0.541 & 0.675 & 0.410 & 0.480 & 0.597 & 0.546 & 0.692 & 0.379 & 0.721 & 0.225 & 0.533 \\
L5 & 0.689 & 0.541 & 0.706 & 0.461 & 0.490 & 0.620 & 0.546 & 0.737 & 0.394 & 0.748 & 0.231 & 0.561 \\
L6 & 0.697 & 0.555 & 0.717 & 0.490 & 0.511 & 0.629 & 0.542 & 0.747 & 0.397 & 0.751 & 0.235 & 0.567 \\
L7 & 0.711 & 0.561 & 0.719 & 0.502 & 0.524 & 0.648 & 0.563 & 0.759 & 0.398 & 0.761 & 0.227 & 0.570 \\
L8 & 0.742 & 0.562 & 0.748 & 0.514 & 0.541 & 0.692 & 0.622 & 0.791 & 0.397 & 0.768 & 0.227 & 0.583 \\
L9 & 0.762 & 0.563 & 0.773 & 0.516 & 0.544 & 0.712 & 0.650 & 0.790 & 0.400 & 0.778 & 0.220 & 0.588 \\
L10 & 0.769 & 0.570 & 0.775 & 0.511 & 0.534 & 0.725 & 0.667 & 0.797 & 0.400 & 0.778 & 0.222 & 0.597 \\
L11 & 0.763 & 0.562 & 0.778 & 0.500 & 0.527 & 0.716 & 0.660 & 0.787 & 0.407 & 0.793 & 0.210 & 0.595 \\
L12 & 0.744 & 0.524 & 0.740 & 0.486 & 0.530 & 0.719 & 0.667 & 0.784 & 0.411 & 0.831 & 0.193 & 0.583 \\ \midrule
\textbf{Average} & \textbf{0.698} & \textbf{0.546} & \textbf{0.703} & \textbf{0.437} & \textbf{0.499} & \textbf{0.639} & \textbf{0.588} & \textbf{0.720} & \textbf{0.383} & \textbf{0.757} & \textbf{0.221} & \textbf{0.553} \\ \midrule
\textbf{Max} & \textbf{0.769} & \textbf{0.570} & \textbf{0.778} & \textbf{0.516} & \textbf{0.544} & \textbf{0.725} & \textbf{0.667} & \textbf{0.797} & \textbf{0.411} & \textbf{0.831} & \textbf{0.235} & \textbf{0.597} \\ \bottomrule
\end{tabular}
}
\caption{\textbf{BERT}: z-score normalization}
\label{tab:intrinsic_task_bert_unit_var}
\end{table*}

\begin{table*}[tbh!]
\centering
\setlength{\tabcolsep}{3pt}
\scalebox{0.65}{
\begin{tabular}{@{}lrrrrrrrrrrrr@{}}
\toprule
\multicolumn{1}{c}{\textbf{L/N}} & \multicolumn{1}{c}{\textbf{MEN}} & \multicolumn{1}{c}{\textbf{WS353}} & \multicolumn{1}{c}{\textbf{WS353R}} & \multicolumn{1}{c}{\textbf{WS353S}} & \multicolumn{1}{c}{\textbf{SimLex999}} & \multicolumn{1}{c}{\textbf{RW}} & \multicolumn{1}{c}{\textbf{RG65}} & \multicolumn{1}{c}{\textbf{MTurk}} & \multicolumn{1}{c}{\textbf{Google}} & \multicolumn{1}{c}{\textbf{MSR}} & \multicolumn{1}{c}{\textbf{SemEval2012\_2}} & \multicolumn{1}{c}{\textbf{Average}} \\ \midrule
L0 & 0.608 & 0.542 & 0.633 & 0.303 & 0.441 & 0.557 & 0.526 & 0.623 & 0.300 & 0.714 & 0.235 & 0.522 \\
L1 & 0.657 & 0.547 & 0.645 & 0.323 & 0.450 & 0.603 & 0.579 & 0.667 & 0.324 & 0.737 & 0.231 & 0.546 \\
L2 & 0.684 & 0.577 & 0.672 & 0.369 & 0.473 & 0.624 & 0.594 & 0.685 & 0.351 & 0.750 & 0.234 & 0.564 \\
L3 & 0.694 & 0.580 & 0.715 & 0.404 & 0.483 & 0.635 & 0.585 & 0.714 & 0.360 & 0.743 & 0.226 & 0.571 \\
L4 & 0.715 & 0.585 & 0.740 & 0.438 & 0.502 & 0.640 & 0.572 & 0.748 & 0.379 & 0.736 & 0.240 & 0.584 \\
L5 & 0.732 & 0.576 & 0.759 & 0.490 & 0.514 & 0.663 & 0.590 & 0.771 & 0.394 & 0.757 & 0.240 & 0.606 \\
L6 & 0.741 & 0.545 & 0.785 & 0.514 & 0.537 & 0.678 & 0.597 & 0.788 & 0.398 & 0.764 & 0.248 & 0.631 \\
L7 & 0.755 & 0.554 & 0.786 & 0.526 & 0.549 & 0.696 & 0.614 & 0.800 & 0.399 & 0.772 & 0.248 & 0.630 \\
L8 & 0.777 & 0.577 & 0.804 & 0.537 & 0.565 & 0.733 & 0.664 & 0.828 & 0.397 & 0.775 & 0.239 & 0.643 \\
L9 & 0.775 & 0.536 & 0.845 & 0.528 & 0.567 & 0.734 & 0.672 & 0.820 & 0.398 & 0.782 & 0.232 & 0.642 \\
L10 & 0.757 & 0.509 & 0.852 & 0.509 & 0.542 & 0.710 & 0.644 & 0.795 & 0.398 & 0.780 & 0.240 & 0.636 \\
L11 & 0.739 & 0.485 & 0.851 & 0.491 & 0.528 & 0.676 & 0.607 & 0.768 & 0.404 & 0.792 & 0.240 & 0.625 \\
L12 & 0.773 & 0.519 & 0.815 & 0.516 & 0.560 & 0.735 & 0.687 & 0.797 & 0.404 & 0.816 & 0.225 & 0.639 \\ \midrule
\textbf{Average} & \textbf{0.723} & \textbf{0.549} & \textbf{0.762} & \textbf{0.458} & \textbf{0.516} & \textbf{0.668} & \textbf{0.610} & \textbf{0.754} & \textbf{0.377} & \textbf{0.763} & \textbf{0.237} & \textbf{0.603} \\ \midrule
\textbf{Max} & \textbf{0.777} & \textbf{0.585} & \textbf{0.852} & \textbf{0.537} & \textbf{0.567} & \textbf{0.735} & \textbf{0.687} & \textbf{0.828} & \textbf{0.404} & \textbf{0.816} & \textbf{0.248} & \textbf{0.643} \\ \bottomrule
\end{tabular}
}
\caption{\textbf{BERT}: all-but-the-top}
\label{tab:intrinsic_task_bert_abtt_raw}
\end{table*}

\begin{table*}[tbh!]
\centering
\setlength{\tabcolsep}{3pt}
\scalebox{0.65}{
\begin{tabular}{@{}lrrrrrrrrrrrr@{}}
\toprule
\textbf{L/N} & \multicolumn{1}{l}{\textbf{MEN}} & \multicolumn{1}{l}{\textbf{WS353}} & \multicolumn{1}{l}{\textbf{WS353R}} & \multicolumn{1}{l}{\textbf{WS353S}} & \multicolumn{1}{l}{\textbf{SimLex999}} & \multicolumn{1}{l}{\textbf{RW}} & \multicolumn{1}{l}{\textbf{RG65}} & \multicolumn{1}{l}{\textbf{MTurk}} & \multicolumn{1}{l}{\textbf{Google}} & \multicolumn{1}{l}{\textbf{MSR}} & \multicolumn{1}{l}{\textbf{SemEval2012\_2}} & \multicolumn{1}{l}{\textbf{Average}} \\ \midrule
L0 & 0.663 & 0.578 & 0.684 & 0.441 & 0.502 & 0.686 & 0.617 & 0.760 & 0.332 & 0.712 & 0.233 & 0.564 \\
L1 & 0.696 & 0.599 & 0.732 & 0.476 & 0.523 & 0.690 & 0.617 & 0.761 & 0.364 & 0.759 & 0.237 & 0.587 \\
L2 & 0.723 & 0.595 & 0.758 & 0.507 & 0.540 & 0.706 & 0.638 & 0.777 & 0.387 & 0.785 & 0.233 & 0.605 \\
L3 & 0.755 & 0.594 & 0.792 & 0.557 & 0.564 & 0.713 & 0.635 & 0.787 & 0.409 & 0.814 & 0.232 & 0.623 \\
L4 & 0.754 & 0.583 & 0.810 & 0.584 & 0.574 & 0.709 & 0.629 & 0.791 & 0.420 & 0.825 & 0.237 & 0.629 \\
L5 & 0.735 & 0.564 & 0.816 & 0.576 & 0.569 & 0.693 & 0.605 & 0.780 & 0.417 & 0.814 & 0.240 & 0.619 \\
L6 & 0.733 & 0.501 & 0.841 & 0.547 & 0.573 & 0.655 & 0.573 & 0.760 & 0.411 & 0.810 & 0.243 & 0.604 \\
L7 & 0.725 & 0.483 & 0.837 & 0.536 & 0.568 & 0.650 & 0.559 & 0.749 & 0.409 & 0.805 & 0.241 & 0.597 \\
L8 & 0.671 & 0.444 & 0.802 & 0.504 & 0.543 & 0.593 & 0.485 & 0.708 & 0.400 & 0.790 & 0.234 & 0.561 \\
L9 & 0.663 & 0.451 & 0.797 & 0.498 & 0.532 & 0.591 & 0.475 & 0.716 & 0.394 & 0.779 & 0.235 & 0.557 \\
L10 & 0.706 & 0.466 & 0.840 & 0.502 & 0.541 & 0.642 & 0.540 & 0.748 & 0.412 & 0.808 & 0.239 & 0.586 \\
L11 & 0.711 & 0.462 & 0.804 & 0.487 & 0.544 & 0.646 & 0.537 & 0.750 & 0.412 & 0.810 & 0.198 & 0.578 \\
L12 & 0.514 & 0.314 & 0.732 & 0.263 & 0.370 & 0.450 & 0.351 & 0.593 & 0.409 & 0.838 & 0.208 & 0.458 \\ \midrule
\textbf{Average} & \textbf{0.696} & \textbf{0.510} & \textbf{0.788} & \textbf{0.498} & \textbf{0.534} & \textbf{0.648} & \textbf{0.559} & \textbf{0.745} & \textbf{0.398} & \textbf{0.796} & \textbf{0.232} & \textbf{0.582} \\ \midrule
\textbf{Max} & \textbf{0.755} & \textbf{0.599} & \textbf{0.841} & \textbf{0.584} & \textbf{0.574} & \textbf{0.713} & \textbf{0.638} & \textbf{0.791} & \textbf{0.420} & \textbf{0.838} & \textbf{0.243} & \textbf{0.629} \\ \bottomrule
\end{tabular}
}
\caption{\textbf{XLNET}: raw embedding}
\label{tab:intrinsic_task_xlnet_raw}
\end{table*}

\begin{table*}[tbh!]
\centering
\setlength{\tabcolsep}{3pt}
\scalebox{0.65}{
\begin{tabular}{@{}lrrrrrrrrrrrr@{}}
\toprule
\multicolumn{1}{c}{\textbf{L/N}} & \multicolumn{1}{c}{\textbf{MEN}} & \multicolumn{1}{c}{\textbf{WS353}} & \multicolumn{1}{c}{\textbf{WS353R}} & \multicolumn{1}{c}{\textbf{WS353S}} & \multicolumn{1}{c}{\textbf{SimLex999}} & \multicolumn{1}{c}{\textbf{RW}} & \multicolumn{1}{c}{\textbf{RG65}} & \multicolumn{1}{c}{\textbf{MTurk}} & \multicolumn{1}{c}{\textbf{Google}} & \multicolumn{1}{c}{\textbf{MSR}} & \multicolumn{1}{c}{\textbf{SemEval2012\_2}} & \multicolumn{1}{c}{\textbf{Average}} \\ \midrule
L0 & 0.665 & 0.555 & 0.702 & 0.407 & 0.495 & 0.678 & 0.628 & 0.750 & 0.334 & 0.711 & 0.227 & 0.559 \\
L1 & 0.691 & 0.562 & 0.728 & 0.448 & 0.515 & 0.691 & 0.636 & 0.765 & 0.367 & 0.762 & 0.241 & 0.582 \\
L2 & 0.714 & 0.565 & 0.734 & 0.489 & 0.531 & 0.699 & 0.641 & 0.774 & 0.389 & 0.788 & 0.232 & 0.596 \\
L3 & 0.760 & 0.582 & 0.779 & 0.543 & 0.557 & 0.714 & 0.646 & 0.795 & 0.412 & 0.812 & 0.230 & 0.621 \\
L4 & 0.760 & 0.582 & 0.795 & 0.568 & 0.571 & 0.707 & 0.634 & 0.792 & 0.420 & 0.824 & 0.234 & 0.626 \\
L5 & 0.744 & 0.582 & 0.784 & 0.564 & 0.570 & 0.693 & 0.612 & 0.782 & 0.417 & 0.815 & 0.239 & 0.618 \\
L6 & 0.741 & 0.584 & 0.790 & 0.560 & 0.568 & 0.691 & 0.614 & 0.785 & 0.412 & 0.807 & 0.244 & 0.618 \\
L7 & 0.745 & 0.583 & 0.795 & 0.557 & 0.562 & 0.695 & 0.625 & 0.782 & 0.409 & 0.802 & 0.241 & 0.618 \\
L8 & 0.719 & 0.571 & 0.758 & 0.538 & 0.538 & 0.659 & 0.569 & 0.765 & 0.405 & 0.792 & 0.231 & 0.595 \\
L9 & 0.717 & 0.568 & 0.756 & 0.534 & 0.541 & 0.663 & 0.574 & 0.768 & 0.403 & 0.785 & 0.230 & 0.594 \\
L10 & 0.758 & 0.583 & 0.816 & 0.546 & 0.554 & 0.700 & 0.623 & 0.792 & 0.417 & 0.812 & 0.238 & 0.622 \\
L11 & 0.777 & 0.583 & 0.810 & 0.552 & 0.565 & 0.733 & 0.661 & 0.809 & 0.420 & 0.829 & 0.231 & 0.634 \\
L12 & 0.798 & 0.562 & 0.864 & 0.553 & 0.563 & 0.767 & 0.702 & 0.829 & 0.430 & 0.859 & 0.231 & 0.651 \\ \midrule
\textbf{Average} & \textbf{0.738} & \textbf{0.574} & \textbf{0.778} & \textbf{0.528} & \textbf{0.548} & \textbf{0.699} & \textbf{0.628} & \textbf{0.784} & \textbf{0.403} & \textbf{0.800} & \textbf{0.235} & \textbf{0.610} \\ \midrule
\textbf{Max} & \textbf{0.798} & \textbf{0.584} & \textbf{0.864} & \textbf{0.568} & \textbf{0.571} & \textbf{0.767} & \textbf{0.702} & \textbf{0.829} & \textbf{0.430} & \textbf{0.859} & \textbf{0.244} & \textbf{0.651} \\ \bottomrule
\end{tabular}
}
\caption{\textbf{XLNET}: z-score normalization}
\label{tab:intrinsic_task_xlnet_unit_var}
\end{table*}

\begin{table*}[tbh!]
\centering
\setlength{\tabcolsep}{3pt}
\scalebox{0.65}{
\begin{tabular}{@{}lrrrrrrrrrrrr@{}}
\toprule
\multicolumn{1}{c}{\textbf{L/N}} & \multicolumn{1}{c}{\textbf{MEN}} & \multicolumn{1}{c}{\textbf{WS353}} & \multicolumn{1}{c}{\textbf{WS353R}} & \multicolumn{1}{c}{\textbf{WS353S}} & \multicolumn{1}{c}{\textbf{SimLex999}} & \multicolumn{1}{c}{\textbf{RW}} & \multicolumn{1}{c}{\textbf{RG65}} & \multicolumn{1}{c}{\textbf{MTurk}} & \multicolumn{1}{c}{\textbf{Google}} & \multicolumn{1}{c}{\textbf{MSR}} & \multicolumn{1}{c}{\textbf{SemEval2012\_2}} & \multicolumn{1}{c}{\textbf{Average}} \\ \midrule
L0 & 0.728 & 0.600 & 0.732 & 0.469 & 0.535 & 0.725 & 0.676 & 0.786 & 0.288 & 0.683 & 0.244 & 0.588 \\
L1 & 0.731 & 0.613 & 0.797 & 0.494 & 0.535 & 0.722 & 0.657 & 0.783 & 0.339 & 0.759 & 0.247 & 0.607 \\
L2 & 0.756 & 0.635 & 0.798 & 0.532 & 0.549 & 0.732 & 0.662 & 0.796 & 0.387 & 0.802 & 0.253 & 0.627 \\
L3 & 0.799 & 0.646 & 0.840 & 0.587 & 0.575 & 0.737 & 0.666 & 0.799 & 0.410 & 0.825 & 0.252 & 0.649 \\
L4 & 0.795 & 0.635 & 0.840 & 0.606 & 0.583 & 0.733 & 0.657 & 0.811 & 0.421 & 0.837 & 0.253 & 0.652 \\
L5 & 0.781 & 0.614 & 0.835 & 0.602 & 0.581 & 0.726 & 0.645 & 0.805 & 0.420 & 0.831 & 0.253 & 0.645 \\
L6 & 0.778 & 0.631 & 0.835 & 0.601 & 0.580 & 0.726 & 0.650 & 0.807 & 0.416 & 0.828 & 0.261 & 0.647 \\
L7 & 0.788 & 0.630 & 0.825 & 0.603 & 0.574 & 0.728 & 0.661 & 0.796 & 0.413 & 0.822 & 0.256 & 0.645 \\
L8 & 0.741 & 0.498 & 0.827 & 0.558 & 0.565 & 0.669 & 0.581 & 0.760 & 0.411 & 0.817 & 0.257 & 0.608 \\
L9 & 0.745 & 0.524 & 0.805 & 0.567 & 0.561 & 0.680 & 0.591 & 0.778 & 0.412 & 0.814 & 0.258 & 0.612 \\
L10 & 0.739 & 0.465 & 0.861 & 0.529 & 0.555 & 0.686 & 0.591 & 0.772 & 0.420 & 0.830 & 0.257 & 0.610 \\
L11 & 0.740 & 0.440 & 0.861 & 0.522 & 0.560 & 0.681 & 0.566 & 0.777 & 0.421 & 0.841 & 0.236 & 0.604 \\
L12 & 0.702 & 0.375 & 0.828 & 0.454 & 0.523 & 0.582 & 0.446 & 0.709 & 0.420 & 0.854 & 0.221 & 0.556 \\ \midrule
\textbf{Average} & \textbf{0.756} & \textbf{0.562} & \textbf{0.822} & \textbf{0.548} & \textbf{0.560} & \textbf{0.702} & \textbf{0.619} & \textbf{0.783} & \textbf{0.398} & \textbf{0.811} & \textbf{0.250} & \textbf{0.619} \\ \midrule
\textbf{Max} & \textbf{0.799} & \textbf{0.646} & \textbf{0.861} & \textbf{0.606} & \textbf{0.583} & \textbf{0.737} & \textbf{0.676} & \textbf{0.811} & \textbf{0.421} & \textbf{0.854} & \textbf{0.261} & \textbf{0.652} \\ \bottomrule
\end{tabular}
}
\caption{\textbf{XLNET}: all-but-the-top}
\label{tab:intrinsic_task_xlnet_abtt}
\end{table*}

\section{Computing Infrastructure and Models' Parameter}
\label{sec:appendix_computing_infras}
We used a server with NVIDIA Tesla V100-SXM2-32 GB GPU, 56 cores, and 500GB CPU memory.

\textbf{Models and Number of Parameters:}
Below, we list the values of the hyper-parameters for different models. 

\begin{itemize}
    \item \textbf{BERT} (bert-base-uncased): L=12, H=768, A=12, total parameters: 110M; where \textit{L} is the number of layers (i.e.,~Transformer blocks), \textit{H} is the hidden size, and \textit{A} is the number of self-attention heads;
    \item \textbf{RoBERTa} (roberta-base): similar to BERT-base, but with a higher number of parameters (125M);
    \item \textbf{XLNet} (xlnet-base-cased) L=12, H=768, A=12, total parameters: 110M.
    \item \textbf{GPT2} L=12, H=768, A=12, total parameters: 117M.

\end{itemize}

\end{document}